\documentclass[10pt,twocolumn,letterpaper]{article}

\usepackage{subfig}
\usepackage{adjustbox}
\usepackage{wacv}
\usepackage{times}
\usepackage{epsfig}
\usepackage{graphicx}
\usepackage{amsmath}
\usepackage{amssymb}
\usepackage{tabularx}
\usepackage{tablefootnote}
\usepackage{booktabs} %
\usepackage{url}

 \usepackage[pagebackref=true,breaklinks=true,letterpaper=true,colorlinks,bookmarks=false]{hyperref}

\wacvfinalcopy %

\def\wacvPaperID{549} %

\newcommand\T{\rule{0pt}{1.2ex}}       %
\newcommand\B{\rule[-0.8ex]{0pt}{0pt}} %

\ifwacvfinal\fi
\begin{document}

\title{Template-Based Automatic Search of Compact Semantic Segmentation Architectures}

\author{Vladimir Nekrasov
\quad
Chunhua Shen
\quad
Ian Reid \\
The University of Adelaide, Australia\\
{
 \small
 E-mail:
 \{vladimir.nekrasov, chunhua.shen, ian.reid\}@adelaide.edu.au 
}
}

\maketitle
\ifwacvfinal\thispagestyle{empty}\fi

\begin{abstract}
	Automatic search of neural architectures for various vision and natural language tasks is becoming a prominent tool as it allows to discover high-performing structures on any dataset of interest. Nevertheless, on more difficult domains, such as dense per-pixel classification, current automatic approaches are limited in their scope -- due to their strong reliance on existing image classifiers they tend to search only for a handful of additional layers with discovered architectures still containing a large number of parameters. In contrast, in this work we propose a novel solution able to find light-weight and accurate segmentation architectures starting from only few blocks of a pre-trained classification network. To this end, we progressively build up a methodology that relies on templates of sets of operations, predicts which template and how many times should be applied at each step, while also generating the connectivity structure and downsampling factors. All these decisions are being made by a recurrent neural network that is rewarded based on the score of the emitted architecture on the holdout set and trained using reinforcement learning. One discovered architecture achieves $63.2\%$ mean IoU on CamVid and $67.8\%$ on CityScapes having only $270$K parameters. Pre-trained models and the search code are available at \url{https://github.com/DrSleep/nas-segm-pytorch}.
	\vskip -0.25in
\end{abstract}

\section{Introduction}

\begin{figure}[t]
	\subfloat[NAS outline\label{fig:nas}]{%
		\begin{minipage}{1.\linewidth}
			\centering
			\includegraphics[width = 0.95\linewidth]{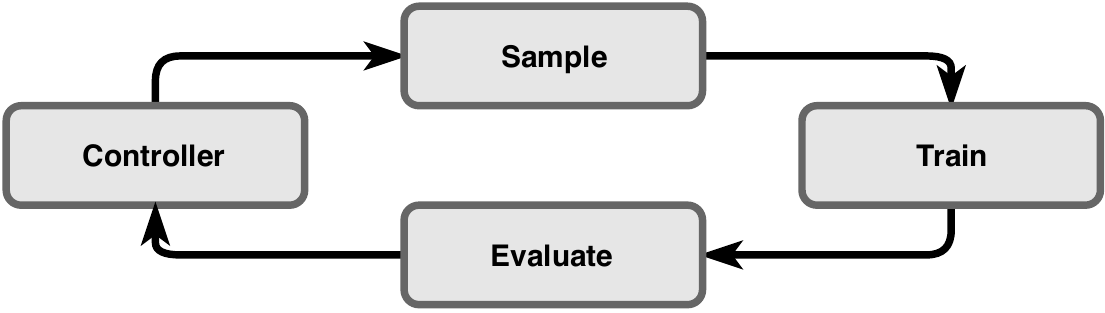}%
		\end{minipage}%
	}
	\hfill
	\hfill
	\subfloat[SOTA comparison\label{fig:csinfo}]{%
		\begin{minipage}{1.\linewidth}
			\centering
			\includegraphics[width = 1.\linewidth]{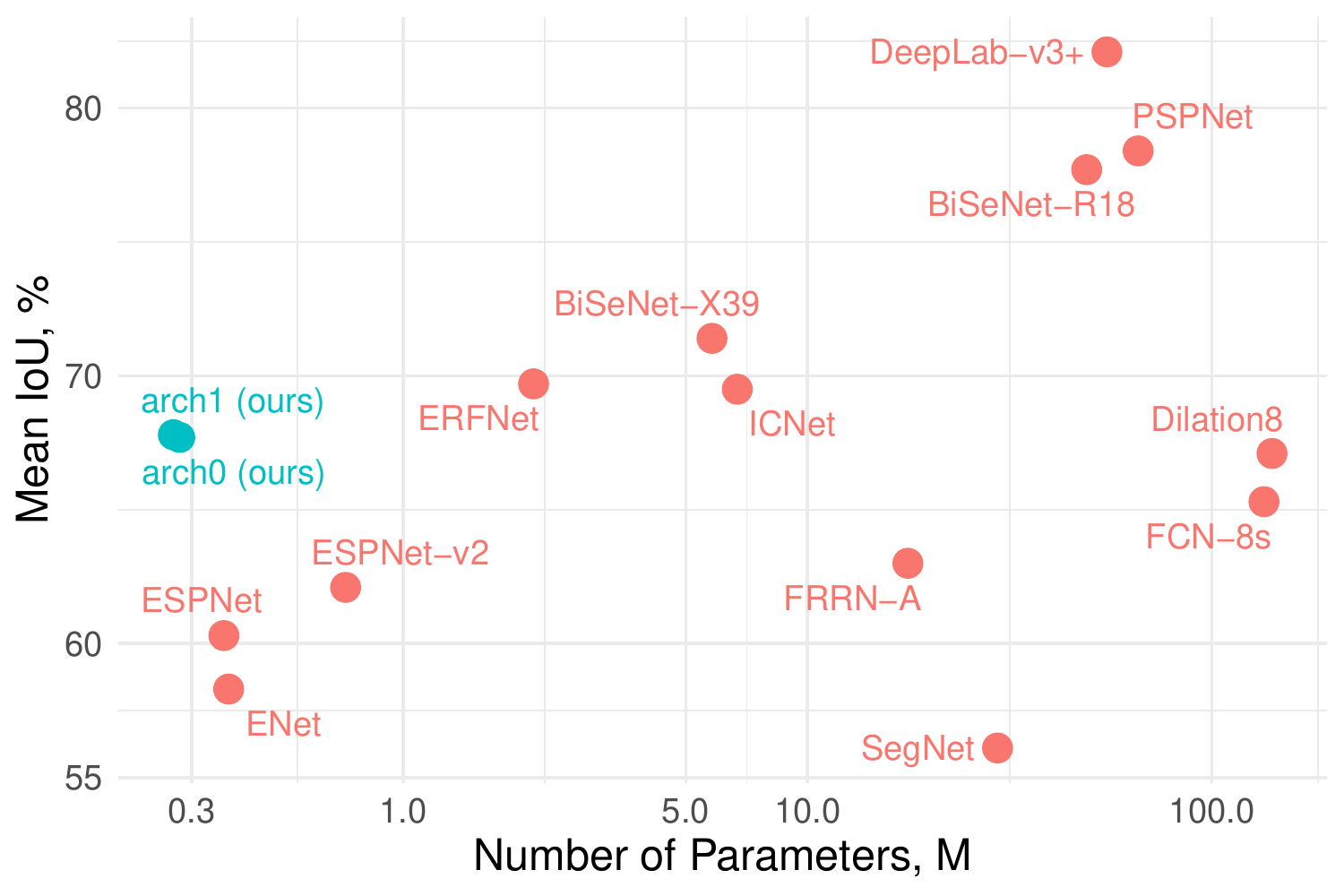}%
		\end{minipage}%
	}
	\caption{({\bf a}) High-level overview of NAS: an architecture is first sampled from a recurrent neural network, controller, and then trained on the meta-train set. Finally, the validation score on the meta-val set is used as the reward signal to train the controller.
		({\bf b}) Comparison of our networks to other methods\protect\footnotemark on the test set of CityScapes~\cite{CordtsORREBFRS16} with respect to the number of parameters and mean IoU.\label{fig:main}}
\end{figure}

While convolutional neural networks (CNNs) keep outperforming competing approaches on various computer vision benchmarks, manually designing novel and more accurate (or more compact) architectures is becoming an increasingly challenging task to handle by human experts. Hence, the recent rise of automatic neural design has turned it into one of appealing solutions in such areas as image classification~\cite{ZophL16}, natural language processing~\cite{ZophVSL17} and even semantic segmentation~\cite{abs-1809-04184}. At its core, the space of thousands of different architectures is traversed with the help of either reinforcement learning~\cite{ZophL16} (Fig.~\ref{fig:nas}), evolutionary strategies~\cite{RealMSSSTLK17} or Bayesian learning~\cite{KandasamyNSPX18}, before a set of `optimal' architectures is found.

~\footnotetext{ENet~\cite{PaszkeCKC16}, ESPNet~\cite{MehtaRCSH18}, ESPNet-v2~\cite{abs-1811-11431}, SegNet~\cite{BadrinarayananK17}, FRRN~\cite{PohlenHML17}, ERFNet~\cite{RomeraABA18}, ICNet~\cite{ZhaoQSSJ18}, FCN-8s~\cite{LongSD15}, Dilation8~\cite{corr/YuK15}, BiSeNet~\cite{YuWPGYS18}, PSPNet~\cite{ZhaoSQWJ17}, DeepLab-v3+~\cite{ChenZPSA18}}

Even for small domains of image classification tasks, this may require an excessive number of resources. For larger domains of dense per-pixel tasks (such as semantic segmentation), where the common practice involves an adaptation of existing image classifiers into fully convolutional networks~\cite{LongSD15}, the neural architecture search ({\em NAS}) methods have been even more limited in their scope. Concretely, recent methods only considered searching for a limited portion of a network and re-used pre-trained image classifiers~\cite{abs-1809-04184,abs-1810-10804}, or used continuous relaxation methodology that severely restricts the search space~\cite{abs-1901-02985}. In situations where the final segmentation network must be extremely light-weight and compact, the classifier-based solution does not fare well as the classifier part requires a significantly larger number of parameters. For example, in the areas of robotics or medical imaging, existing datasets tend to comprise only a scarce set of annotations, and, as evident by recent successes (\eg U-Net~\cite{RonnebergerFB15}), compact networks may well be surprisingly sufficient in such domains.

A naive approach of overcoming the aforementioned limitation may first include the search of compact classification networks followed by the search of operations specific for semantic segmentation. While potentially working, it would require a significant amount of resources to carry out two instantiations of NAS, and might be sub-optimal as the structures found for image classification may still possess redundant operations that are not necessary for semantic segmentation. Another way might rely on the direct search of segmentation architectures end-to-end: reinforcement learning-based existing methodologies will not fare well as they would require to make an exceedingly large number of sequential decisions, while continuous relaxation-based methods will severely limit the search space and may not explore enough architectures. Hence we face a difficult challenge of how to compactly represent the space of segmentation architectures in a way that is representative enough and not over-complicated for the search algorithm to solve.

In our approach, we are motivated by the so-called `cell' (or `motif') strategies~\cite{ZophVSL17,abs-1711-00436}, where a sequence of operations is tied together to form a template. Starting from a tiny stem of a pre-trained classifier, the search algorithm generates several such templates together with the full structure of the segmentation network. At each decision step, it predicts two locations of layers to be used in a template, the template number, the number of repetitions that the template will be used for and the downsampling factor (stride) of the first operations in the template.

As we show in the experimental part, such an approach leads to flexible representations of end-to-end architectures for semantic segmentation. We search on CityScapes~\cite{CordtsORREBFRS16}, and train best discovered architectures both on it and another common benchmark, CamVid~\cite{BrostowFC09}. Our smallest model with only $270$K parameters achieves $67.8\%$ and $63.2\%$ mean IoU, on CityScapes and CamVid, respectively, which compares favourably to other methods (Fig.~\ref{fig:csinfo}).
It must also be noted that our methodology requires only $2$ GPUs to carry out the search process.

In conclusion to this introduction, we re-iterate that our first and foremost contribution is to showcase a simple method for searching semantic segmentation architectures with a limited reliance on only few layers of a pre-trained image classifier.  We methodically build up our solution and approach the issues defined above in the following sections.

\section{Related Work}

\subsection*{Semantic Segmentation}

In semantic segmentation the goal is to come up with a per-pixel semantic labelling of the given image. Over the last years, most prominent approaches have been built upon fully convolutional networks~\cite{LongSD15}, where image classifier's fully-connected layers are being converted into convolutional ones. Important and task-specific design choices include skip-connections~\cite{LongSD15}, dilated convolutions~\cite{ChenPKMY18} and contextual blocks~\cite{ChenPKMY18, LinMSR17, ZhaoSQWJ17} -- all of which tend to improve semantic segmentation results.

Recently, several manually designed segmentation architectures have emerged that achieved highly accurate results across common benchmarks with compact networks. In particular, Romera~\etal~\cite{RomeraABA18} altered the popular residual block~\cite{HeZRS16} by replacing $3{\times}3$ convolutions in it with factorised counterparts --~\ie $3{\times}1$ and $1{\times}3$ convolutions. On the test set of CityScapes~\cite{CordtsORREBFRS16} their model, ERFNet, achieved $69.7\%$ mean IoU having $2.1$M parameters. Guided by the same principle of efficiently and effectively enlarging the receptive field size, Mehta~\etal~\cite{MehtaRCSH18} replaced a convolutional layer with a hierarchical pyramid of point-wise and dilated convolutions. On the same benchmark, the proposed approach, ESPNet, showed $60.3\%$ mean IoU with only $0.4$M parameters. Later, Mehta~\etal~\cite{abs-1811-11431} made ESPNet even more efficient by exploiting separable convolutions -- ESPNet-v2 with $0.7$M parameters attained $62.1\%$ mean IoU. The authors of BiSeNet~\cite{YuWPGYS18} relied on two paths instead: one that keeps the spatial information intact, and one contextual with aggressive downsampling strategy. Two paths are merged by a specifically designed fusion module with per-channel attention. Their model obtained $71.4\%$ mean IoU with $5.8$M parameters.

While these manually designed approaches do show promising performance, coming up with even better models is becoming extremely challenging for human experts. In contrast, our solution is automatic and able to find high-performing compact architectures.

\subsection*{NAS background}

We rely on NAS using reinforcement learning (RL)~\cite{ZophVSL17}, where a recurrent neural network (`controller') sequentially outputs actions that when fused together fully define an architecture. The emitted architecture is then trained on the task of interest, and its validation score is sent back to the controller as a reward signal. The controller is trained with proximal policy optimisation (PPO)~\cite{SchulmanWDRK17} to maximise the average reward of emitted architectures.

The first NAS works in image classification~\cite{ZophL16} emitted a full description of the layers to use -- for example, the number of input / output channels, kernel sizes, strides, etc. Exploring the space with all those decisions required excessive resources, hence, an alternative solution was proposed, where the search algorithm is tasked with discovering only two sets of operations -- so-called, \emph{reduction} and \emph{normal} cells~\cite{ZophVSL17}, that are stacked and arranged in blocks to form the final network. Naturally, this limits the search space but tends to work well in practice. In a similar vein, Liu~\etal~\cite{abs-1711-00436} defined so-called \emph{motifs} that are used inside an evolutionary search algorithm and undergo mutations depending on their fitness scores. To speed-up the search process and reduce the overload, several methods considered to instantiate all possible architectures beforehand and either choose the single path with RL~\cite{PhamGZLD18} or perform weighted average with continuous relaxation~\cite{abs-1806-09055}. Inherently, this line of work considers only a limited set of architectures and is able to explore only the paths pre-defined in the beginning of the search process.

NAS approaches for semantic segmentation have followed in the footsteps of NAS in image classification -- in particular, Chen~\etal~\cite{abs-1809-04184} searched for a single cell on top of a fully convolutional classifier, while Nekrasov~\etal~\cite{abs-1810-10804} searched for the decoder part inside the encoder-decoder type of the segmentation network. Both methods extensively rely on existing image classifiers and thus are limited in the way of describing the search space. Besides, even though in~\cite{abs-1810-10804} the goal is to find compact segmentation networks, their lower bound in terms of efficiency is already pre-defined by the choice of a classifier. In contrast, our search setup allows to discover extremely tiny networks without relying on expensive tricks such as knowledge distillation and Polyak averaging as in~\cite{abs-1810-10804}. Furthermore, our method no longer requires a complete encoder and instead finds architectures starting only from $60$K pre-trained parameters.

Most recently, Liu~\etal~\cite{abs-1901-02985} also adapted the continuous learning relaxation for semantic segmentation and achieved impressive results. As mentioned above, such methodology has very low diversity as the space of architectures to explore is confined. While future advances may well overcome these limitations, in this work we instead concentrate on the RL-based approach.
\section{Our methodology}

As the stem of our architecture we consider three initial residual blocks of MobileNet-v2~\cite{abs-1801-04381} -- in total they contain only $60$K parameters and reduce the spatial resolution to $\frac{1}{8}$. This aggressive downsampling in the beginning is a common feature among most fully convolutional networks. Notably, we generate a significantly larger portion of the network automatically.%

From the stem, we record two outputs of $\frac{1}{4}$ and $\frac{1}{8}$ spatial resolution. Given those, for the rest of the network we use a recurrent neural network, controller, to predict actions $a^{i}_{j}$ at each step $i$ of block $j$. We begin from the sequence of actions used in~\cite{abs-1810-10804} that looks as follows: $A=[loc_{1}, loc_{2}, op_{1}, op_{2}]$, where $loc_{i}$ is the layer the operation $op_{i}$ will be applied to.\footnote{From here on we omit the block indices and use capital letters to denote the sequence of tokens.}

In Sect.~\ref{htm} we start by describing our template modelling approach motivated by cells and motifs, in Sect.~\ref{ints} we propose another modification that allows us to control the depth of the network, and in Sect.~\ref{strides} we discuss how to alter the spatial resolution.

\subsection{Hierarchical template modelling}
\label{htm}

\begin{figure}[t]
	\begin{center}
		\includegraphics[width=1.\linewidth]{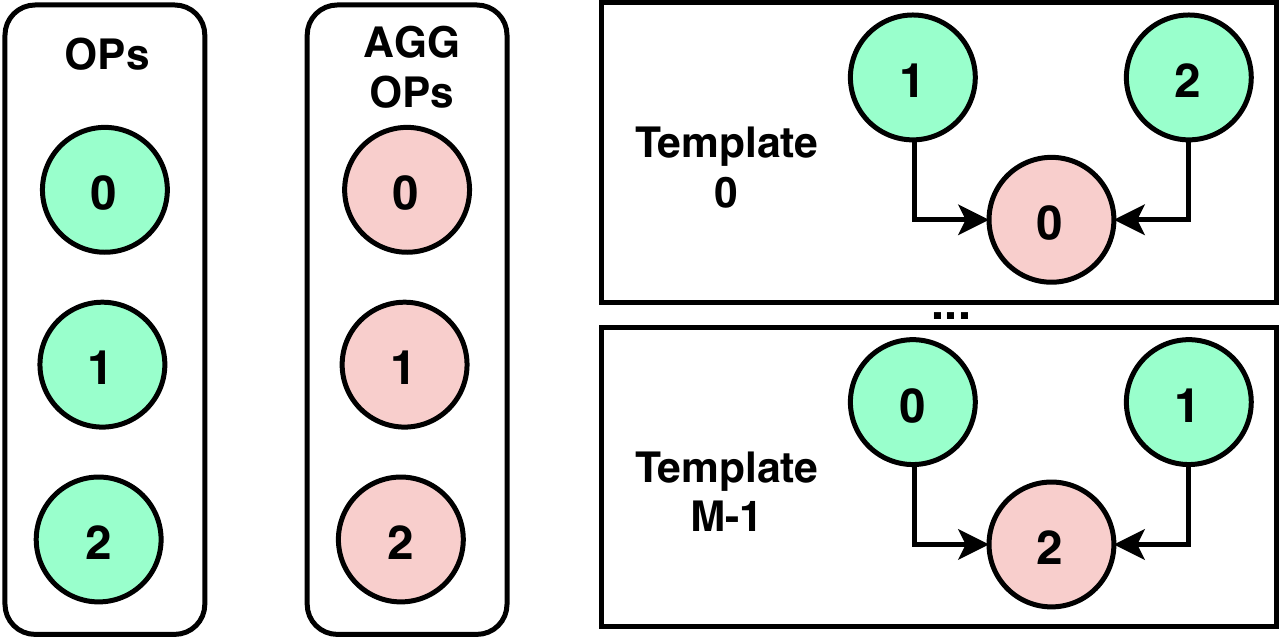}
	\end{center}
	\caption{We generate $M$ templates each of which takes two inputs and produces one output. The template comprises two individual operations and one aggregation operation.\label{fig:template}}
\end{figure}

First of all, we note that the definition of the action sequence above is restrictive: if we were to apply the same arrangement of operations somewhere in the network again, we would need to sample it again, too -- which would be wasteful. To this end, we separate the sampling process of locations and operations. In particular, we generate a set of templates of operations that can be plugged into the network at multiple locations. Inside the template, we also include the aggregation operation $op_{agg}$ (to be either per-pixel summation or channel-wise concatenation). The template takes two inputs and produces one output. Each input undergoes the corresponding operation, and the aggregation operation is used on the intermediate outputs~(Fig.~\ref{fig:template}). Thus, any template can be written as follows: $T = [op_{1}, op_{2}, op_{agg}]$. In cases where inputs have unequal number of channels, the output channel dimension of each operation is set to the largest among the inputs.

\begin{figure}[t]
	\begin{center}
		\includegraphics[width=1.\linewidth]{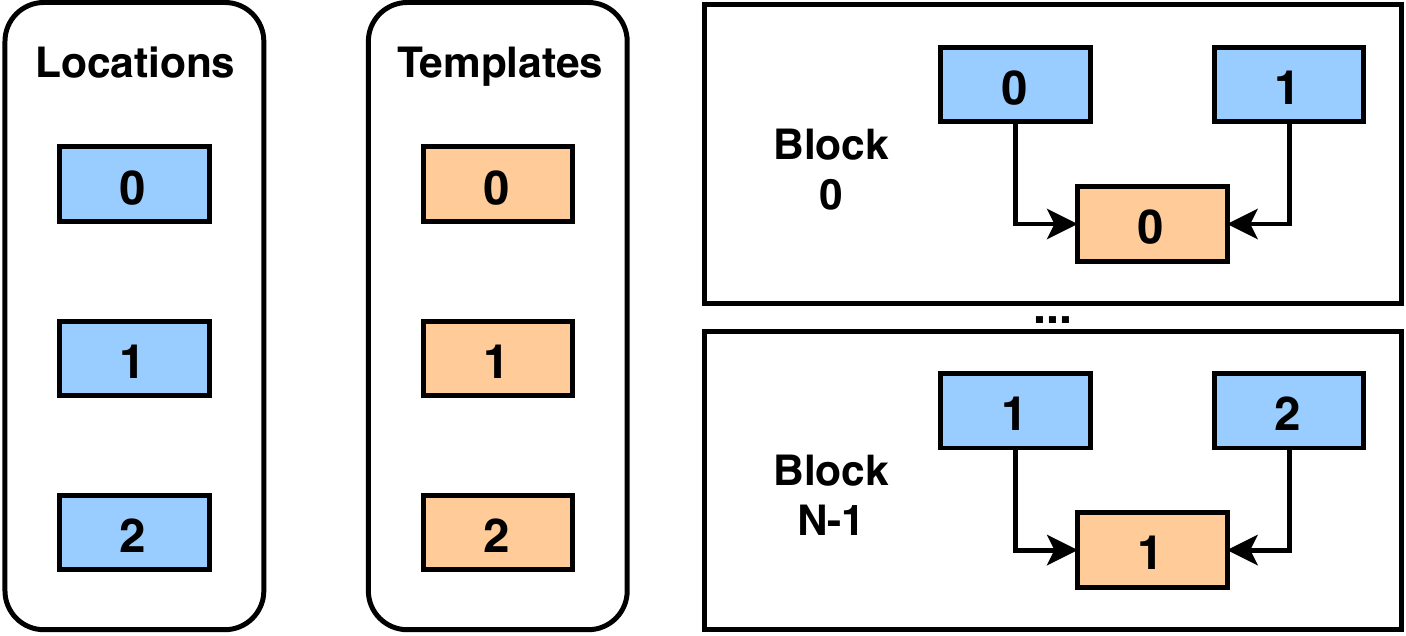}
	\end{center}
	\caption{We generate $N$ blocks by sampling two locations and one template applied to them.
		\label{fig:structure}}
\end{figure}

Having generated the templates, we move on to generating the network structure: in particular, we sample (with replacement) two locations out of the sampling pool (initialised with two stem outputs), and the index of the template~(Fig.~\ref{fig:structure}) -- $A=[loc_{1}, loc_{2}, id_{T}]$. The template's output is added into the sampling pool and the process is repeated multiple times. In the end, we concatenate all non-sampled outputs from the sampling pool, reduce their channel dimension with the help of $1{\times}1$ convolution and predict per-pixel labels with a single $3{\times}3$ convolutional layer.

The benefit of using templates in this approach is that the number of decisions to be made increases slowly with the number of blocks: \eg consider the length of the sentence that describes the architecture -- it would be $(2 + 3) * N$ for the baseline solution, where $N$ is the number of blocks, and $(2 + 1) * N + 3 * M$ for the template one, where $M$ is the number of templates. Having the number of templates lower than $2/3$ of the number of blocks would require less decisions to be made for the same network depth. At the same time, the template modelling would allow the controller to efficiently re-use existing designs.

\subsection{Increasing the number of templates}
\label{ints}

While the template modelling has its benefits, it still possesses one significant disadvantage: in order to generate a deeper network, we can only increase the number of blocks, which, in turn, would lead to significantly more decisions to be made. Hence, we propose to include an additional parameter in the structure generator at the cost of having $N$ more decisions to be made. Concretely, we generate $k$ -- the number of times the template must be repeated -- we consider $k$ to take values from $1$ to $4$; accordingly, the action sequence then becomes $A=[loc_{1}, loc_{2}, id_{T}, k]$. While we could have also abstracted away $k$ into the template definition to reduce the number of decisions, we chose not to because at different blocks it might be optimal to vary $k$.

\begin{figure}[t]
	\begin{center}
		\includegraphics[width=1.\linewidth]{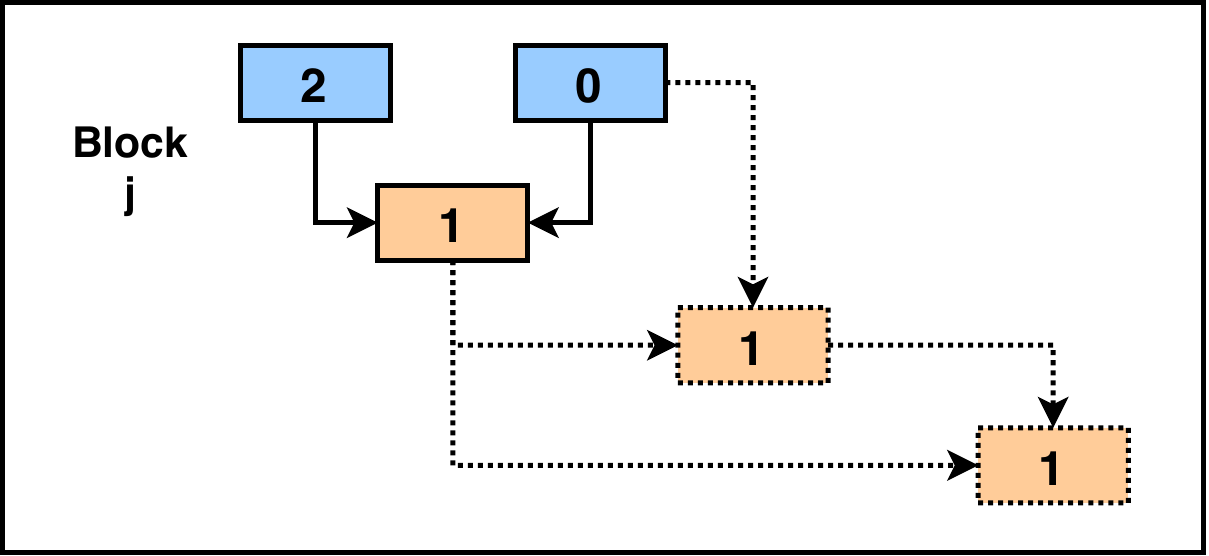}
	\end{center}
	\caption{A template can be recursively applied multiple times (with non-shared weights): the output of the previous template becomes the first input to the current one, and the final output is considered as the block's output. New instantiations of the template together with connections are depicted with dotted lines. 
		\label{fig:repeats}}
\end{figure}

The process of repeating the template is as follows: after the template is applied on two sampled locations and the template's output is recorded, the second input and the template's output are considered as another two inputs to a new instantiation of the same template but with different weights~(Fig.~\ref{fig:repeats}). This is repeated $k$ times, and the last output is appended to the sampling pool.

\subsection{Adding strides}
\label{strides}

It must be noted that up until now we did not make any assumptions with regards to the strides of operations used with the exception of the fixed stem block. Nevertheless, in order to generate a complete segmentation architecture it is important to consider the downsampling factors. Keeping all consecutive operations at a constant stride would not fare well as the common wisdom suggests that in order to extract better features, we need to keep reducing the spatial dimensions while increasing the number of channels until a certain point.

Taking this into account, we append an additional decision to make in the sequence definition: the stride prediction -- either $1$ or $2$. If the stride is 2, we decrease spatial dimensions and multiply the channel dimension by a pre-defined constant that would allow us to control the compactness of the generated network if needed. We only predict strides for the first half of all the blocks, and assume that the rest of them uses stride $1$. Analogously, to deal with inputs of varying resolutions when applying an aggregation operation, we downsample the inputs to the lowest resolution among them for the first half, and upsample to the largest resolution -- for the rest. This is similar to the encoder-decoder architecture design that fares well in semantic segmentation.
The stride prediction is taken out of the template definition to allow the same template to be used with varying strides at different locations. Likewise, it is straightforward to add the prediction of dilation rates per template to recover the dilated decoder approach~\cite{ChenPKMY18}, but it is out of scope of this paper.

Our final string describing a complete architecture can be written as follows:
${[loc_{1}, loc_{2}, id_{T}, k, s]}{\times}N$, where $N$ is the number of blocks, $T=[op_{1}, op_{2}, op_{agg}]$, and the number of templates is $M$. As can be easily seen, we decomposed the original decision sequence into multiple components that allowed us to be compact and flexible.

\section{Search Experiments}

\subsection{NAS setup}
\label{nas:setup}
As the stem of the network, we use three first blocks of MobileNet-v2~\cite{abs-1801-04381} pre-trained on ImageNet~\cite{RussakovskyDSKS15}; two outputs from the second and third blocks with $24$ and $32$ channels and the spatial resolution of $\frac{1}{4}$ and $\frac{1}{8}$ from the original size, respectively, are added into the initial sampling pool. During search, the number of channels is doubled after spatial downsampling. In the beginning, all layers in the sampling pool are transformed to have $48$ channels with the help of $1{\times}1$ convolution. The pre-classifier layer has the same number of $48$ channels.

To keep the number of potential architectures to discover at reasonable level, we set the number of templates and the number of layers to $3$ and $7$, correspondingly. The maximum number of times the template can be used sequentially is set to $4$.

As the search dataset, we consider the training split of CityScapes~\cite{CordtsORREBFRS16} with $2975$ images randomly divided into meta-train ($2677$, or $90\%$) and meta-val ($298$). We resize all images to have a longer side of $1024$ and train on square crops of $321{\times}321$. Each sampled architecture is trained for $10$ epochs and validated twice every $5$ epochs. For the first five epochs, we pre-compute stem outputs and only train the generated part of the network; for the second five, the whole network is trained end-to-end. As the reward we employ the geometric mean of mean IoU, mean accuracy and frequency-weighted IoU as done in~\cite{abs-1810-10804}. %
To speed up the convergence of sampled architectures, we rely on Adam with the learning rate of $7{e-}3$, used on mini-batches of $32$ examples.

We consider the following set of operations
\begin{itemize}
	\itemsep -.122cm
	\item separable conv $3\times3$,
	\item separable conv $5\times5$,
	\item global average pooling followed by upsampling and conv $1\times1$,
	\item max-pool $3\times3$,
	\item separable conv $5\times5$ with dilation rate $6$,
	\item skip-connection.
\end{itemize}
There are also two aggregation operations -- summation and concatenation.

To train the controller, we use PPO~\cite{SchulmanWDRK17} with the learning rate of $0.0001$. In total, we sample, train and evaluate over $2000$ architectures in $8$ days using two $1080$Ti GPUs.

\subsection{Analysis of Search Results}

\begin{figure}[thb]
	\begin{center}
		\includegraphics[width=1.\linewidth]{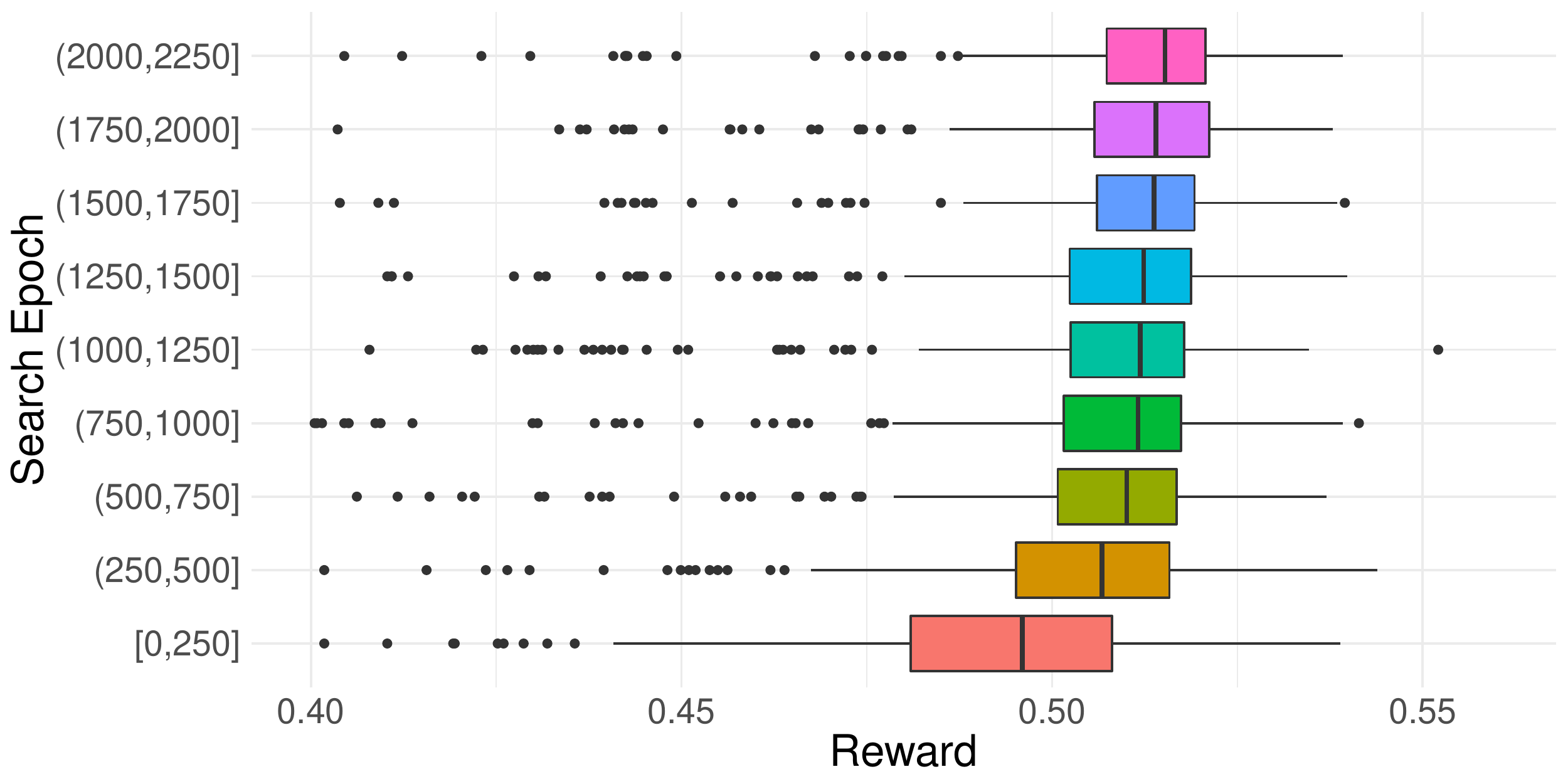}
	\end{center}
	\vskip -0.2in
	\caption{Distribution of rewards attained by architectures sampled by the controller. For compactness of the plot, we only visualise rewards greater than or equal to $0.40$.\label{fig:rewards}}
	\vskip -0.2in
\end{figure}

We first study the rewards progress through time. The median reward is steadily increasing with more epochs as can be inferred from Fig.~\ref{fig:rewards}. While most architectures are tightly clustered together, there are several notable outliers on the far right with the reward of near $0.55$ that we will explore in full training experiments in Sect.~\ref{sec:training}.%

\begin{figure}[thb]
	\begin{center}
		\includegraphics[width=1.\linewidth]{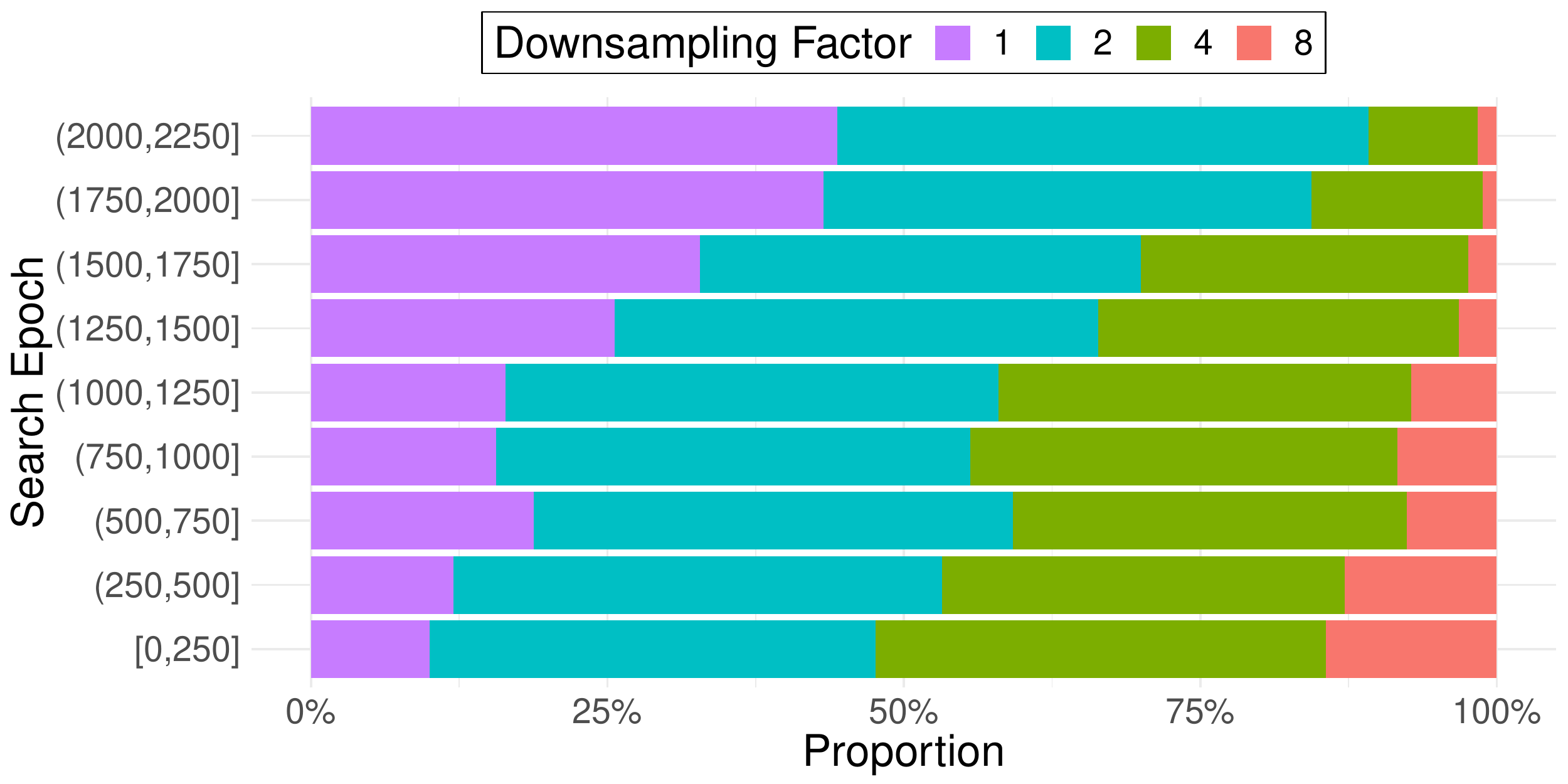}
	\end{center}
	\vskip -0.2in
	\caption{Proportion of downsampling factors through time. The minimum downsampling of $1$ happens when the controller chooses to use stride=1 everywhere; the maximum downsampling of $8$ happens when the controller uses stride=2 three times (hence, $2^{3}=8$).\label{fig:strides}}
\end{figure}

We also consider how the resolution of the architecture is related to its reward score. To this end, we first analyse how often the controller chose to downsample\footnote{With $7$ blocks the maximum number of times downsampling can be performed is $\lfloor{7/2}\rfloor=3$. Note that here we do not take into account the stem resolution.}. Considering that all outcomes are equally likely in the beginning, one would expect the extreme resolutions of $1$ and $\frac{1}{8}$ to be less present. As seen on Fig.~\ref{fig:strides}, it is indeed what happens with the controller at the start of the training. Nevertheless, by the end of the search process the controller becomes far too conservative with regards to downsampling and is more likely to pass on actions that reduce the spatial dimensions. This raises a question of how the downsampling factor in the generated architecture relates to the reward?

\begin{figure}[thb]
	\begin{center}
		\includegraphics[width=1.\linewidth]{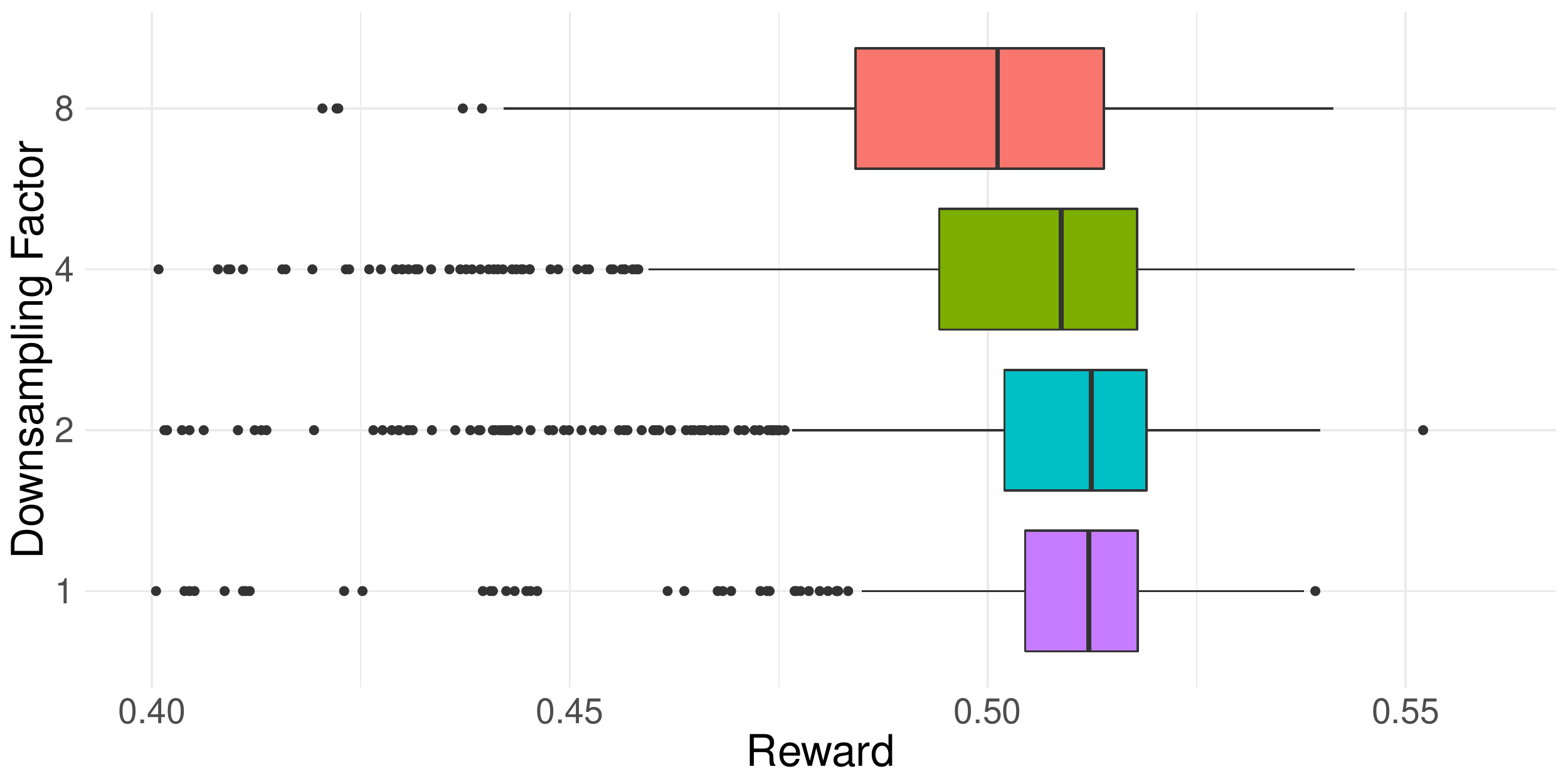}
	\end{center}
	\vskip -0.2in
	\caption{
		Distribution of rewards attained by architectures sampled by the controller with varying downsampling factors. For compactness of the plot, we only visualise rewards greater than or equal to $0.40$.\label{fig:strides:rewards}}
\end{figure}

In an attempt to answer that question, we visualise the reward distribution of architectures with different downsampling factors on Fig.~\ref{fig:strides:rewards}. As the plot implies, the rewards for architectures with higher resolution tend to be larger, hence, the controller becomes biased towards sampling fewer downsampling actions.

Please refer to our supplementary material for more analysis of the search results.%

\subsection{Comparison with Random Search}

An important question to ask with NAS is whether the trained controller performs reliably better than a naive baseline --~\eg random search. Relevant to that is the question of whether the ranking of architectures based on rewards achieved during the search is well-correlated with the ranking of architectures based on their scores during a longer training.

We strive to answer those questions by performing an experiment similar to the one in~\cite{abs-1810-10804}: concretely, we sample two sets with $20$ architectures each - the first set is coming from the pre-trained controller, while the second set is sampled randomly. Each set is trained and evaluated on two setups - the search one as described in Sect.~\ref{nas:setup} and the one where the training continues for more epochs (we increase it to $20$ for the second stage).

\begin{figure}[thb]
	\begin{center}
		\includegraphics[width=1.\linewidth]{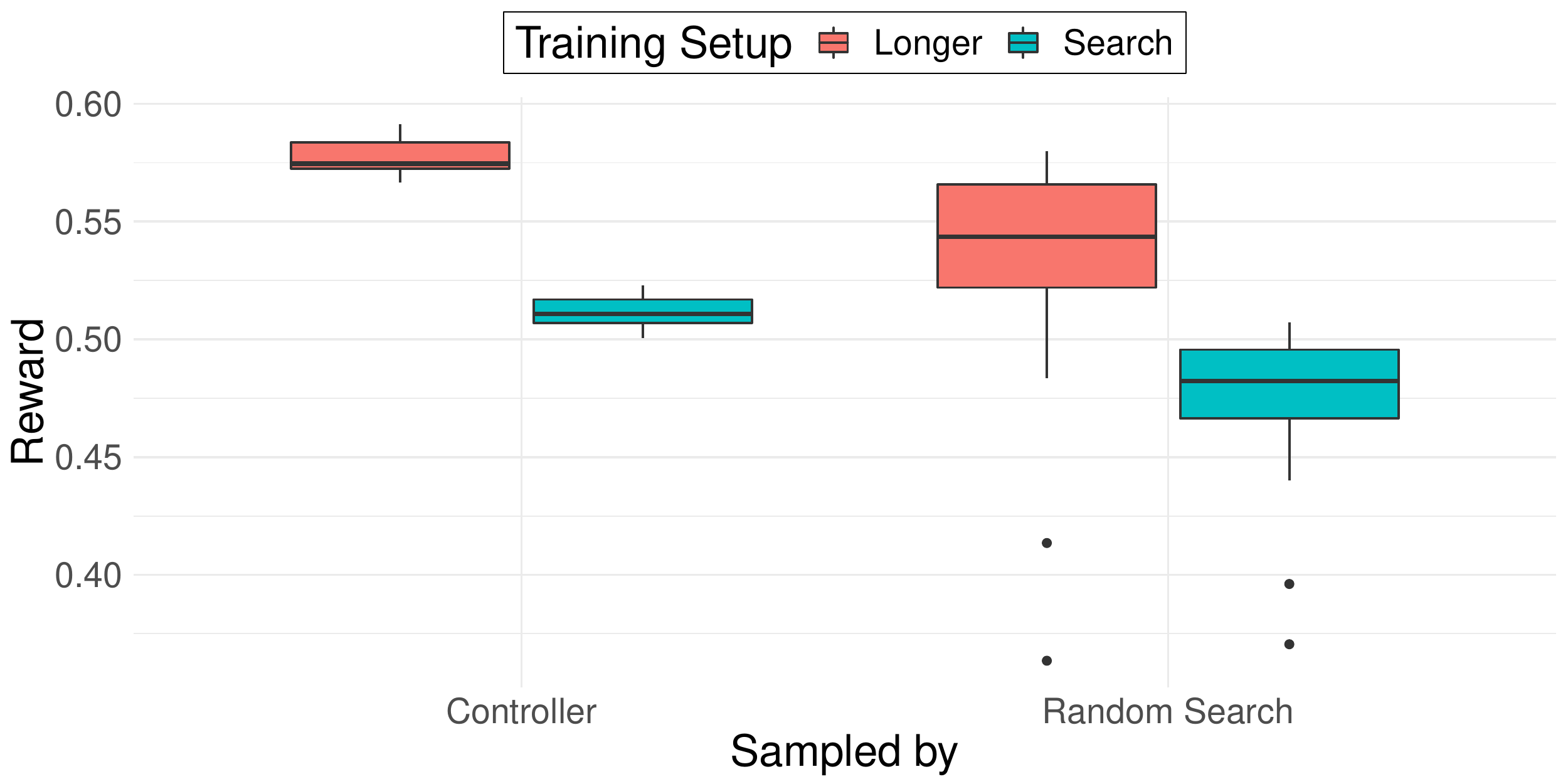}
	\end{center}
	\vskip -0.2in
	\caption{Rewards distribution of $40$ architectures, $20$ of which are sampled by the trained controller and $20$ by random search.\label{fig:rlrs}}
\end{figure}

We visualise the performance results in Fig.~\ref{fig:rlrs}. The architectures sampled by the trained controller reliably achieve higher rewards - both within the search and longer training setups.

\begin{figure}[thb]
	\begin{center}
		\includegraphics[width=1.\linewidth]{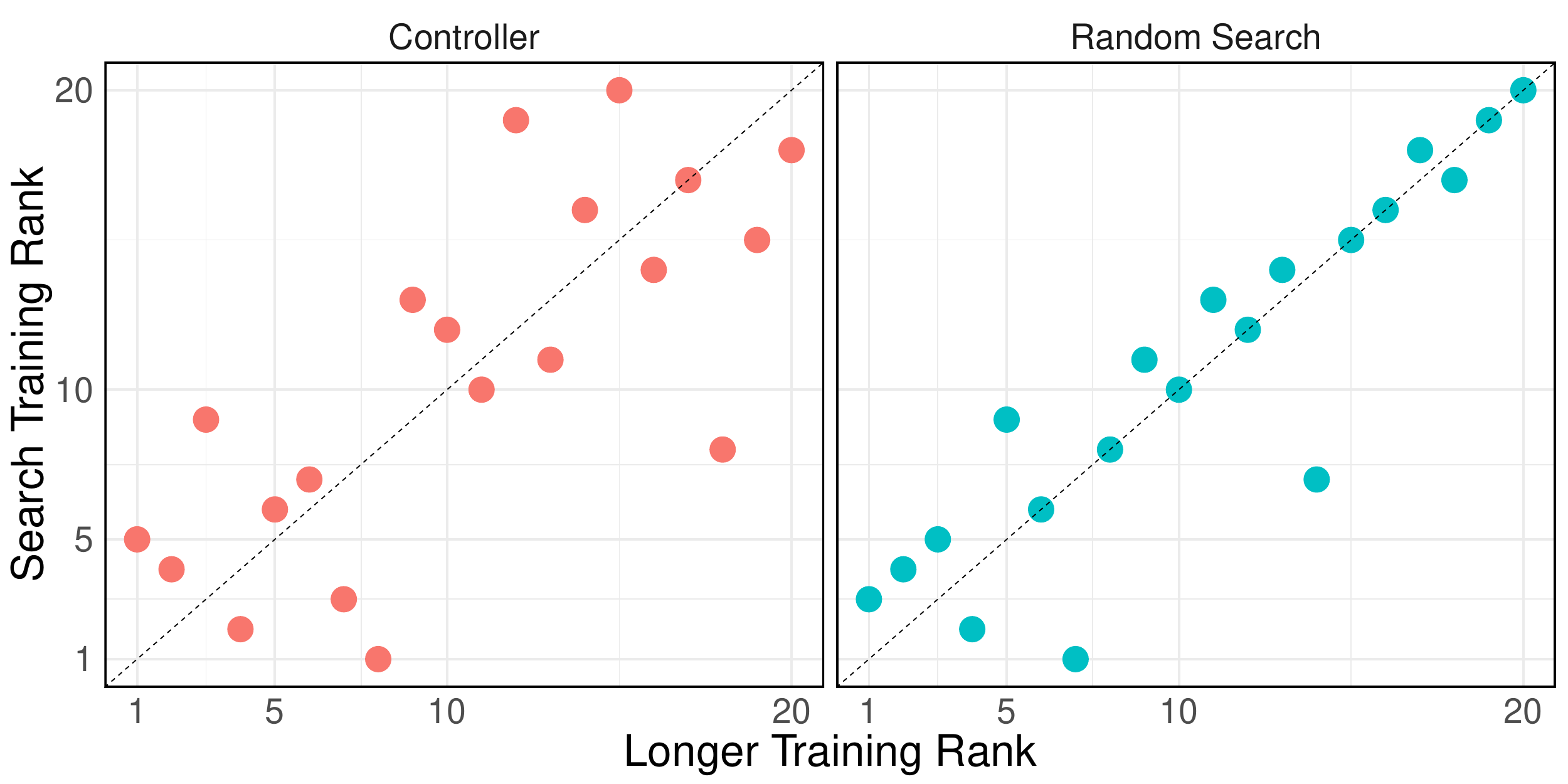}
	\end{center}
	\vskip -0.2in
	\caption{Ranks of architectures based on rewards during the longer training setup ($x$-axis) and the search training setup ($y$-axis).\label{fig:ranks}}
\end{figure}

We further plot corresponding rankings on Fig.~\ref{fig:ranks}. As evident from the plot, when trained for longer the architectures tend to be ranked similarly to the order attained during the search process. In particular, high values of Spearman's rank correlation -- $\rho=0.734$ for the controller, and $\rho=0.904$ for random search -- indicate that the rewards achieved during the searching process may serve as a reliable estimate of the architecture's potential.

\begin{figure*}[t]
	\centering
	\resizebox{1.\textwidth}{!}{\begin{tabular}{cc|cc}
			\subfloat{\includegraphics[width = 0.19\linewidth]{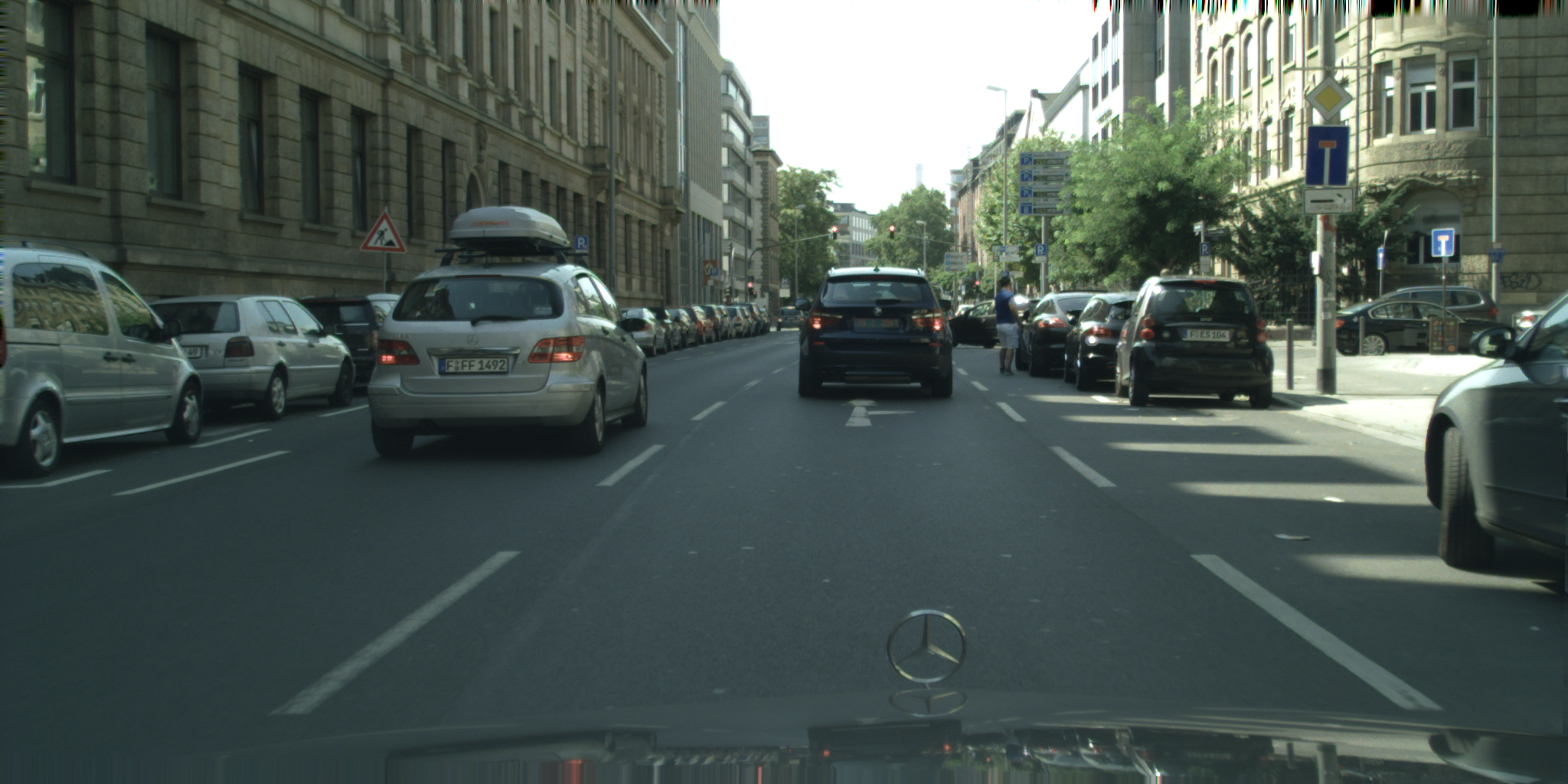}} &
			\subfloat{\includegraphics[width = 0.19\linewidth]{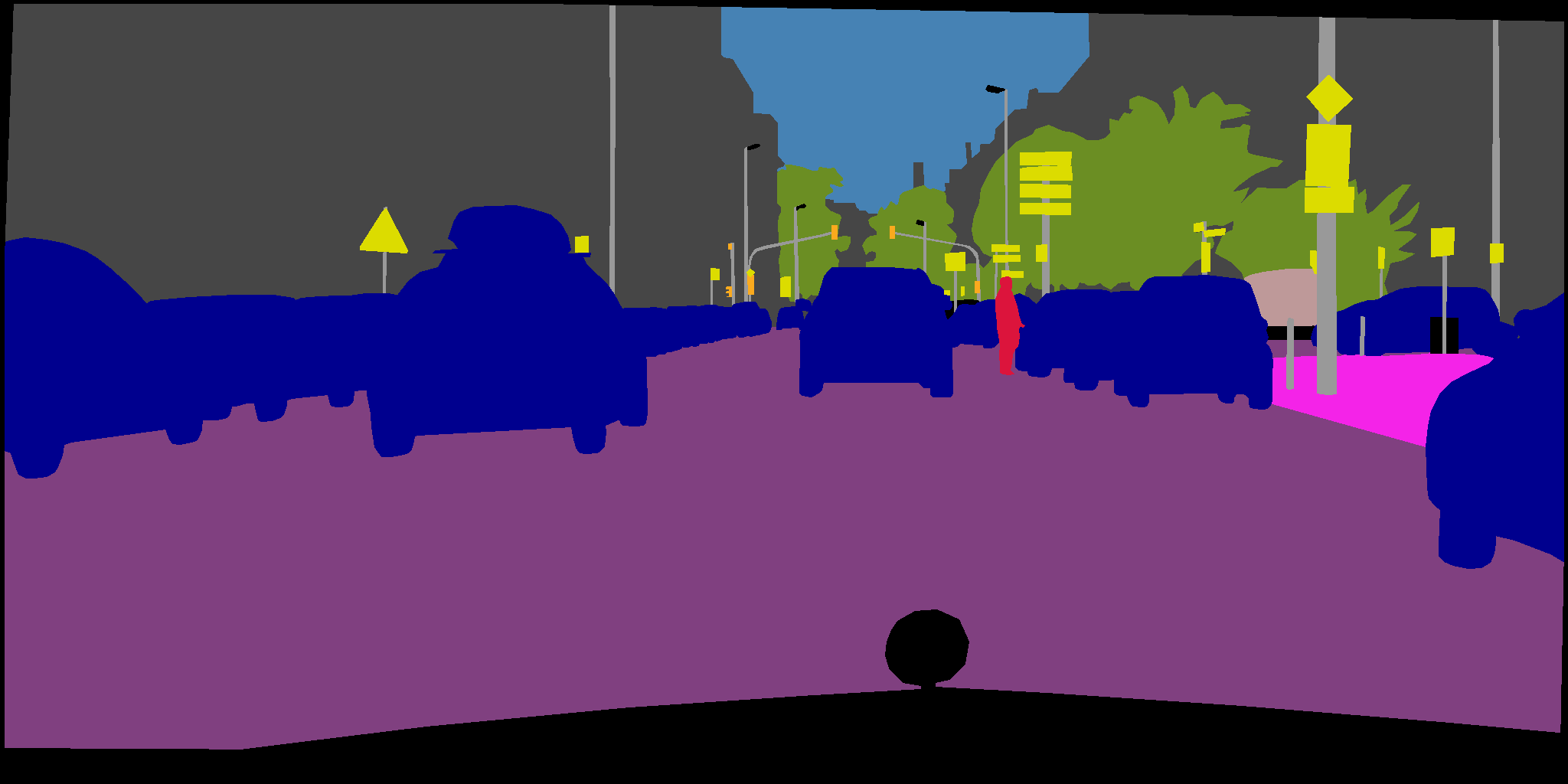}} &
			\subfloat{\includegraphics[width = 0.19\linewidth]{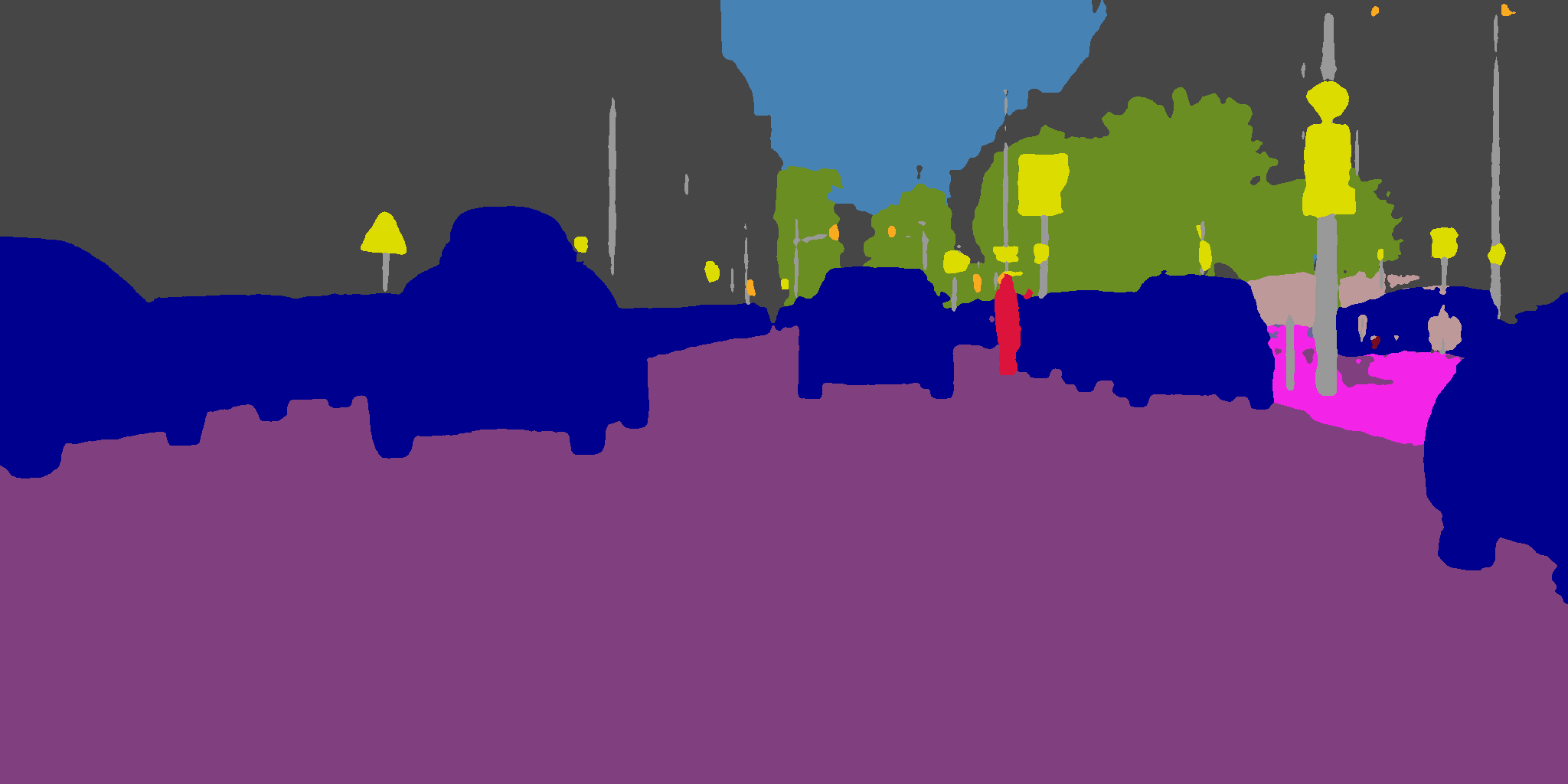}} &
			\subfloat{\includegraphics[width = 0.19\linewidth]{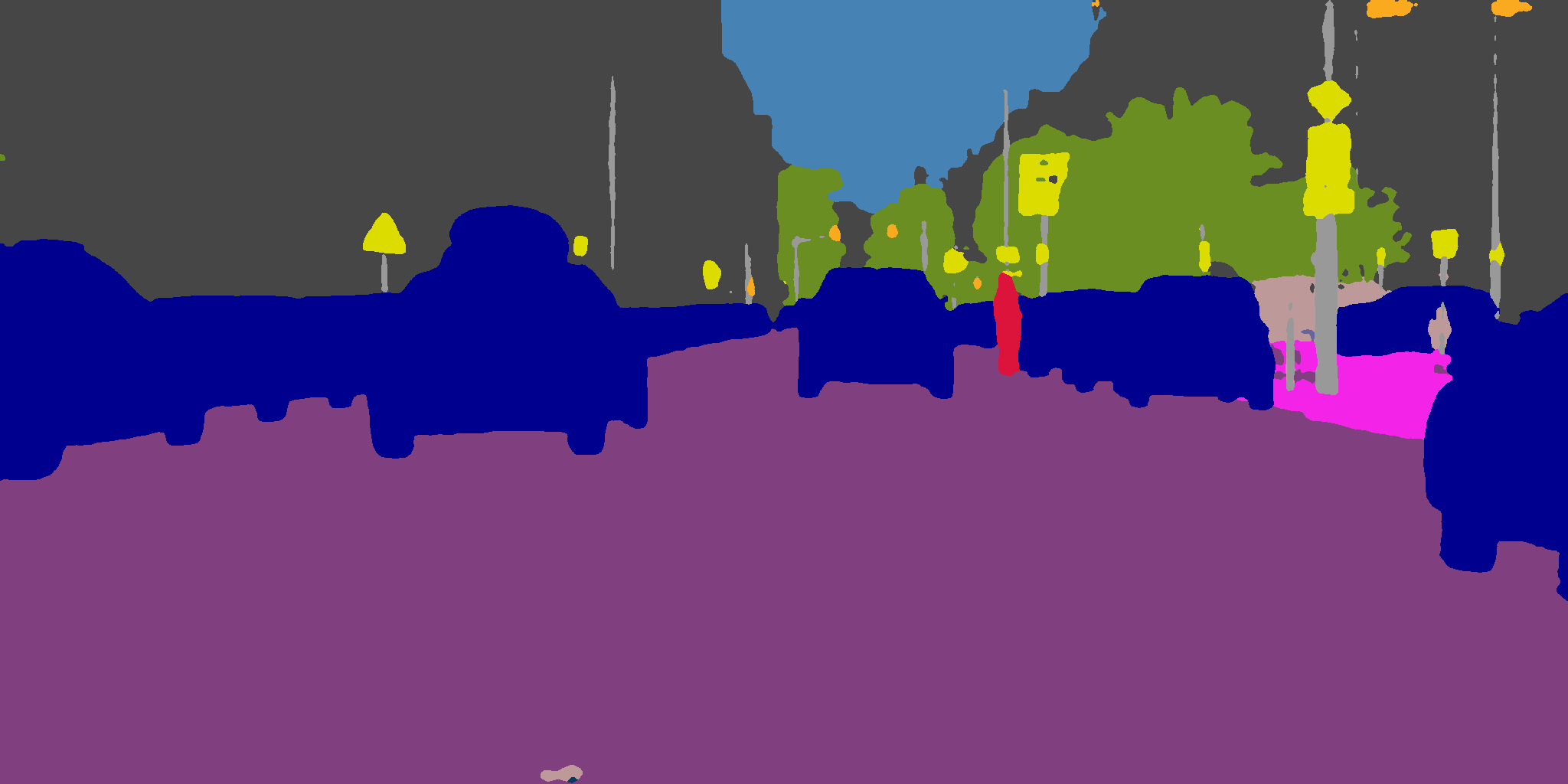}}\\[-0.15in]
			\subfloat{\includegraphics[width = 0.19\linewidth]{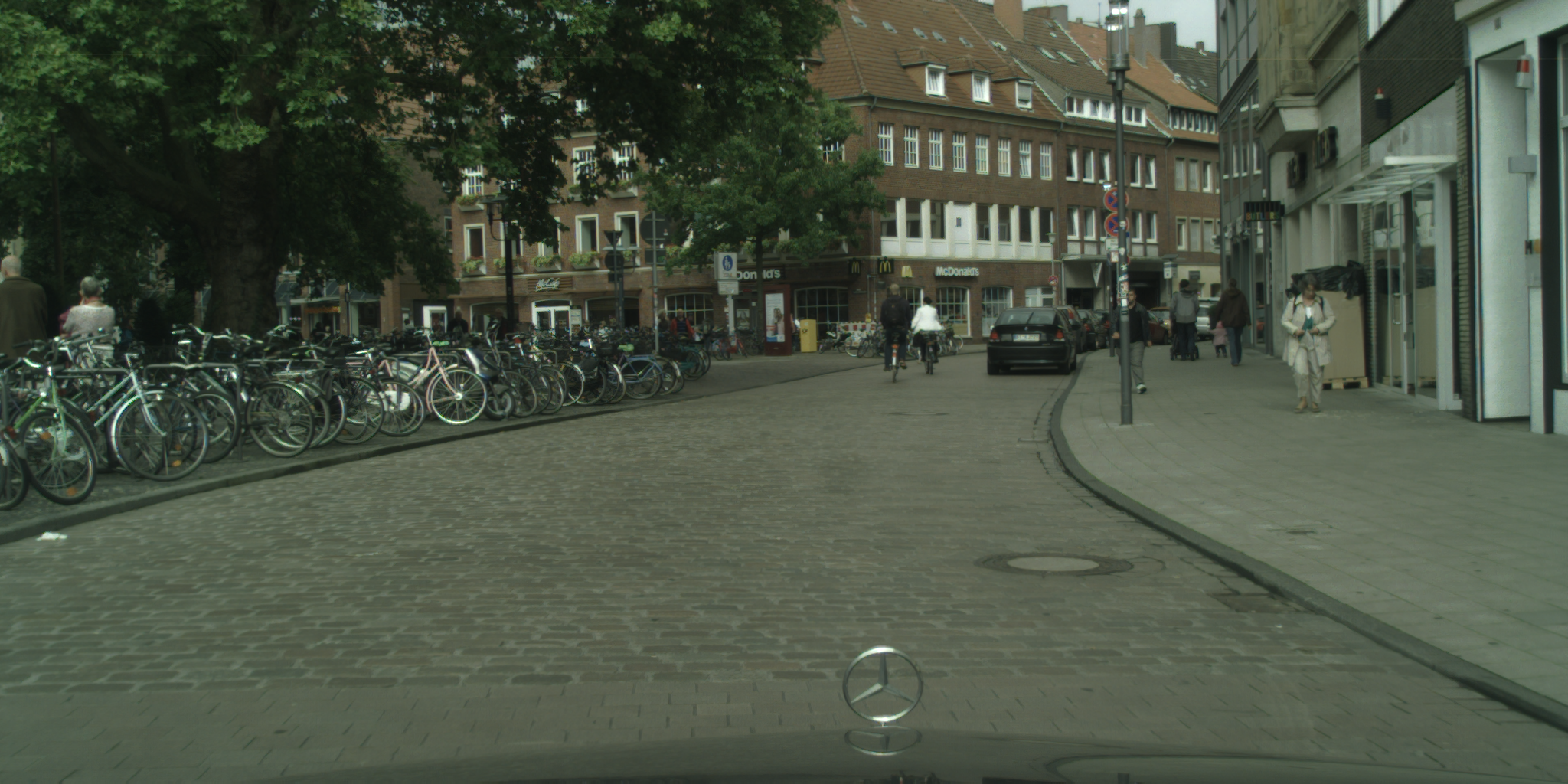}} &
			\subfloat{\includegraphics[width = 0.19\linewidth]{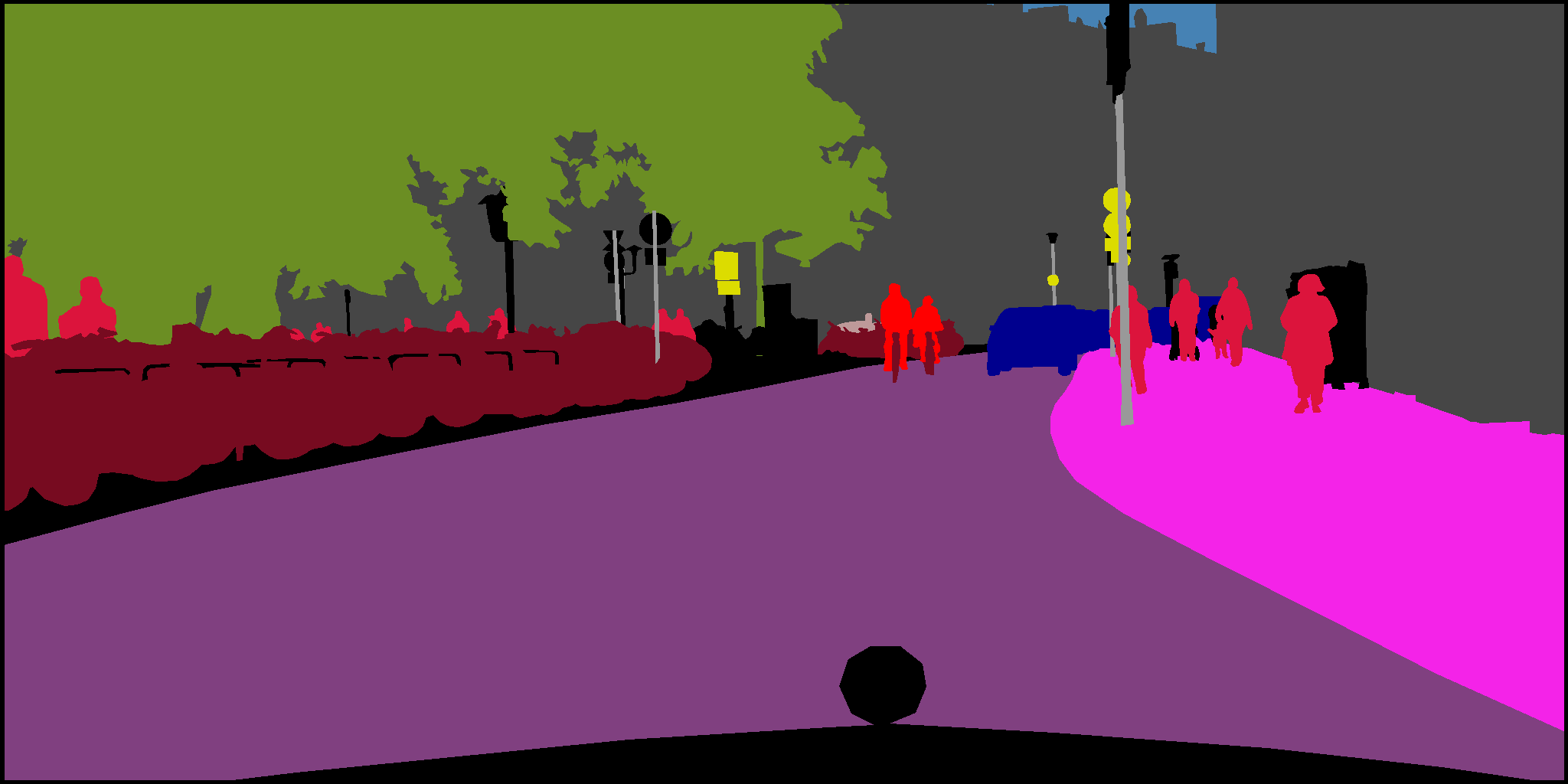}} &
			\subfloat{\includegraphics[width = 0.19\linewidth]{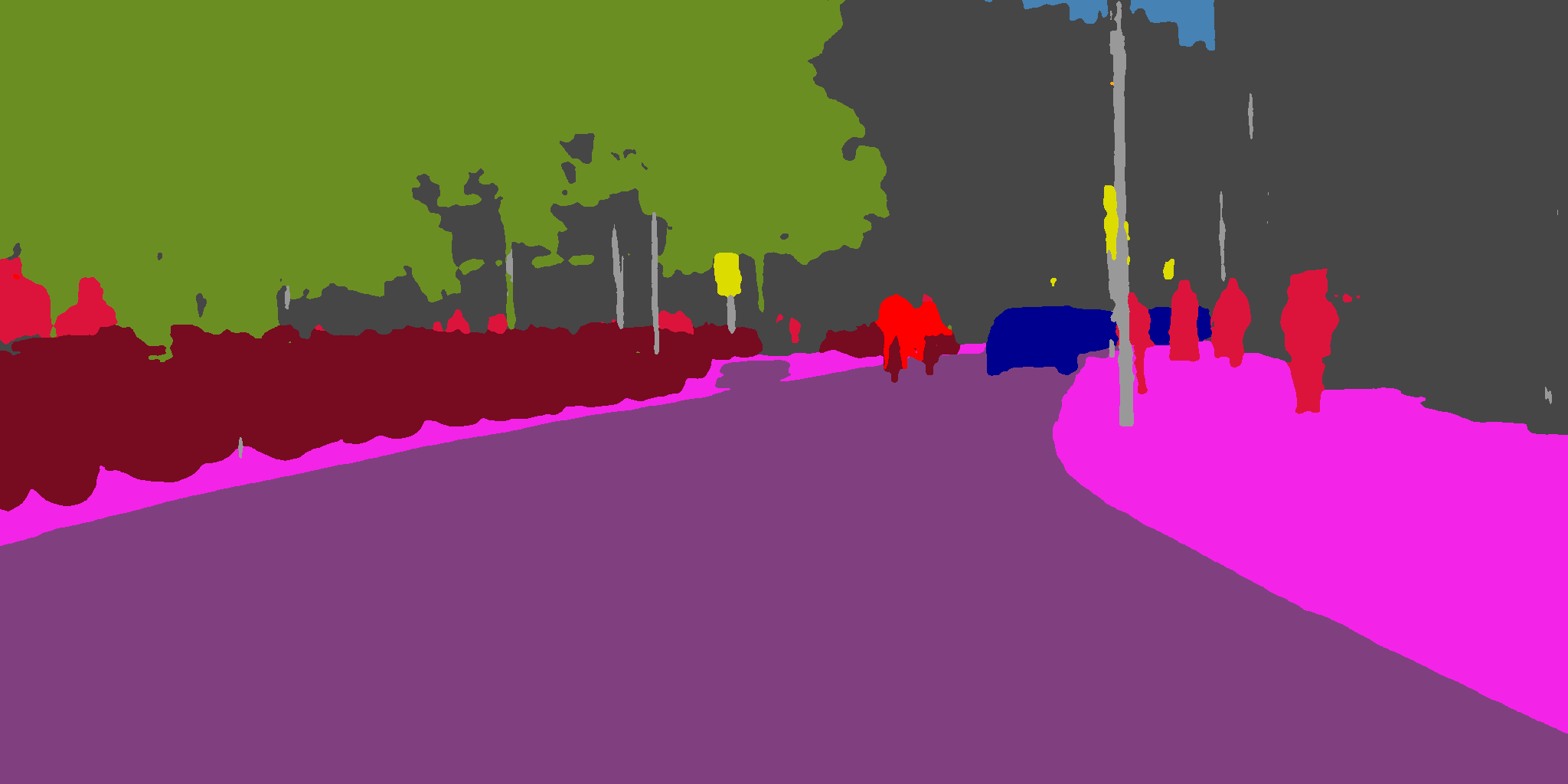}} &
			\subfloat{\includegraphics[width = 0.19\linewidth]{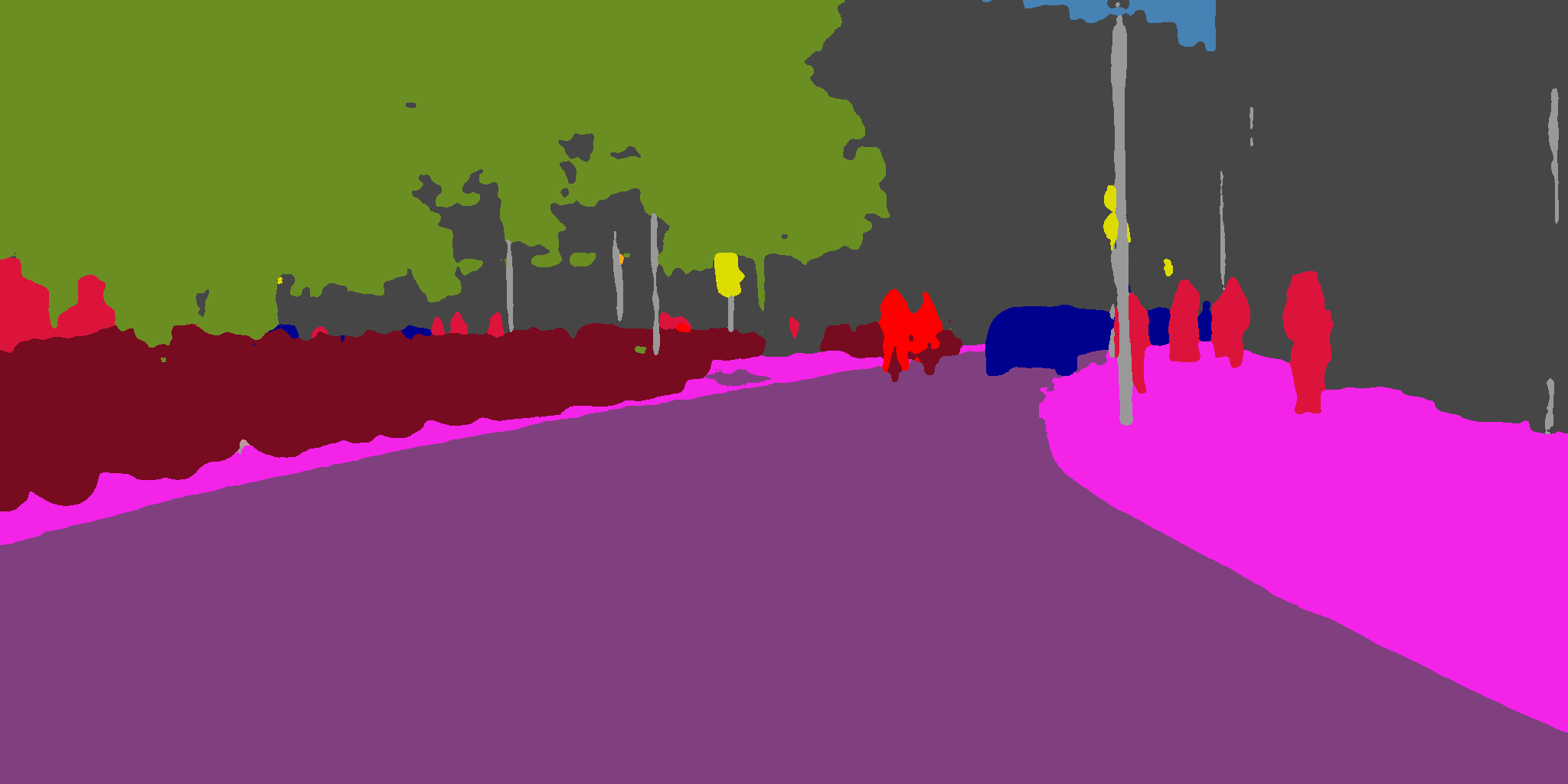}}\\[-0.15in]
			\subfloat{\includegraphics[width = 0.19\linewidth]{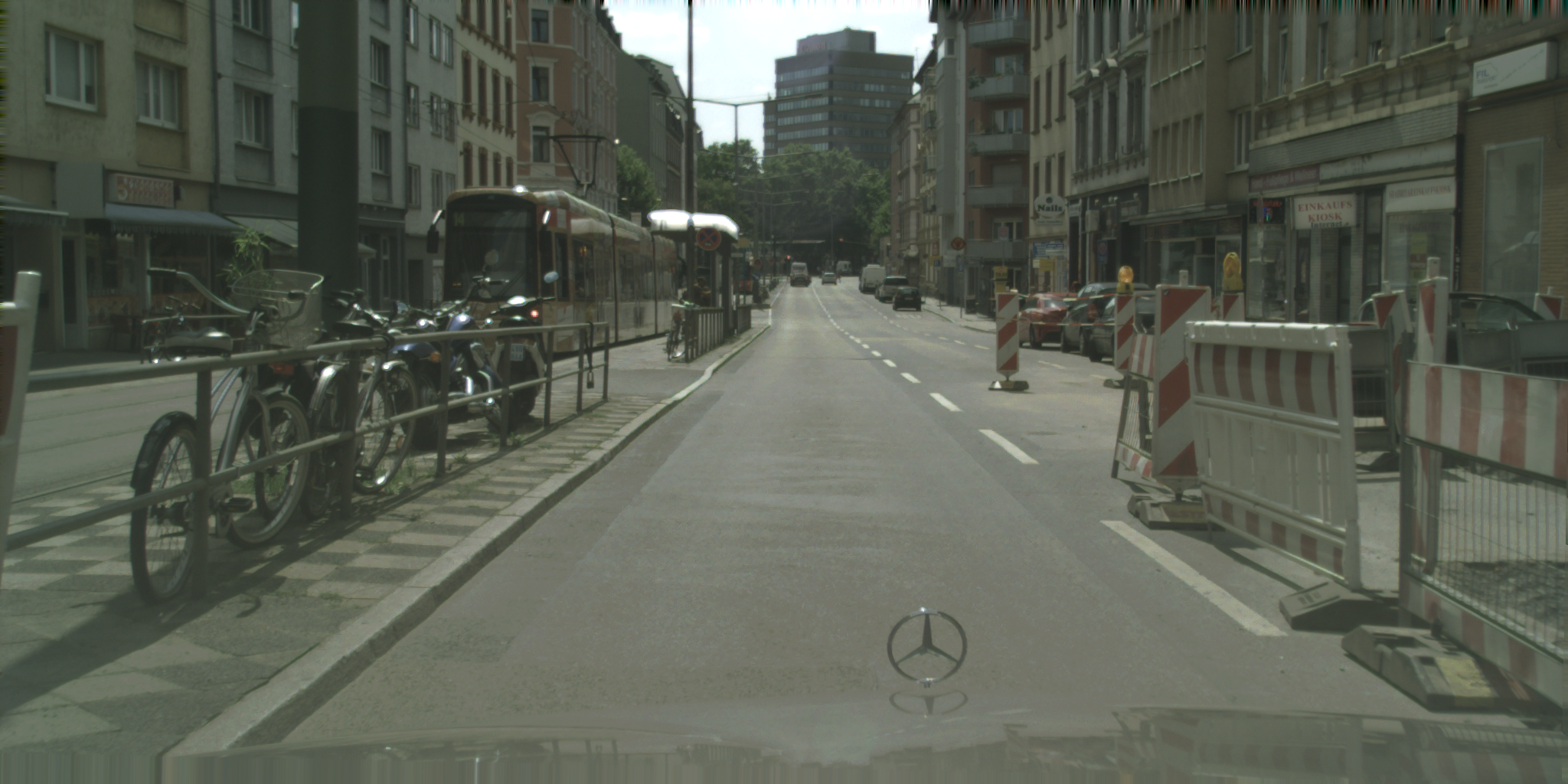}} &
			\subfloat{\includegraphics[width = 0.19\linewidth]{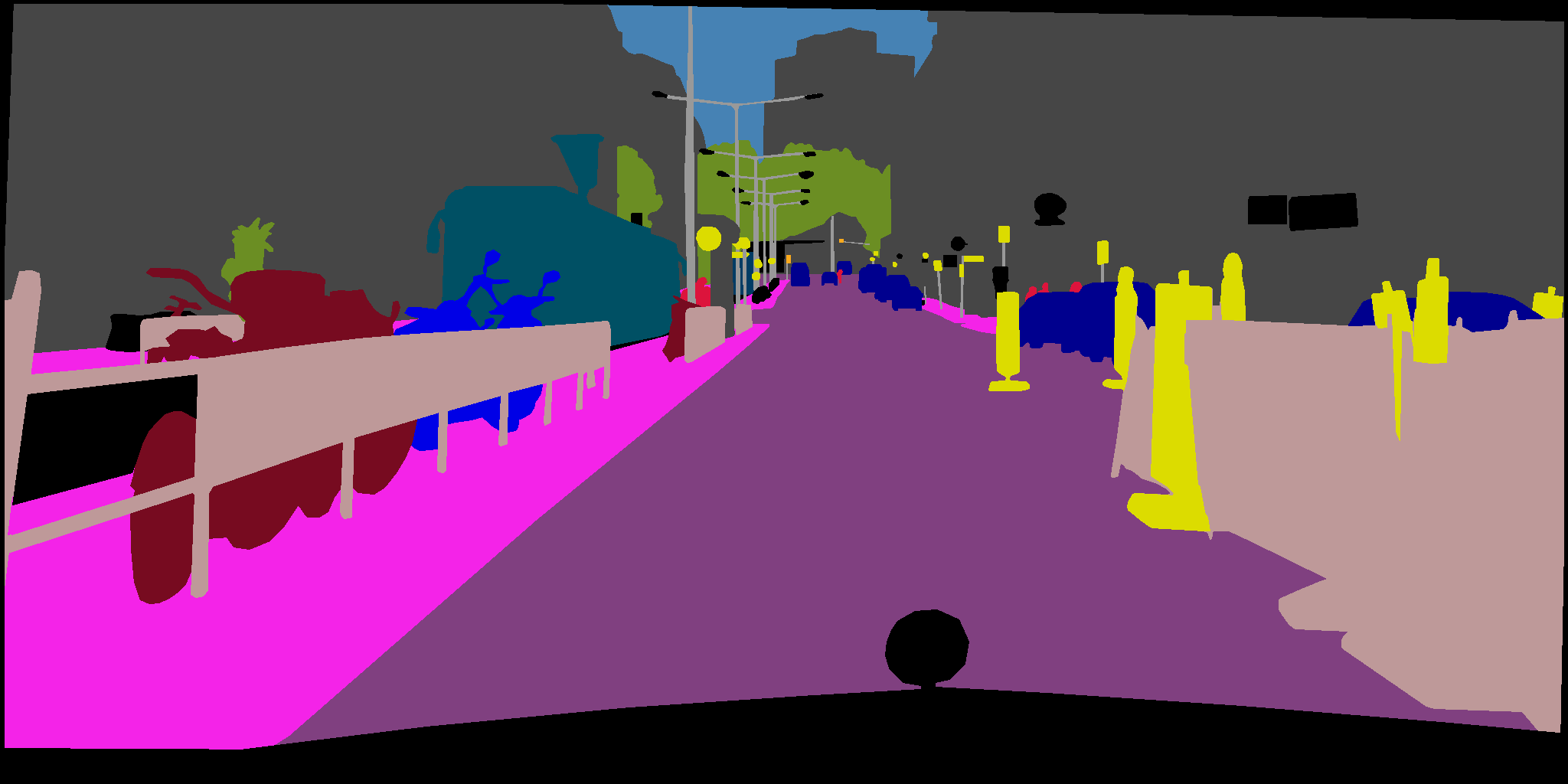}} &
			\subfloat{\includegraphics[width = 0.19\linewidth]{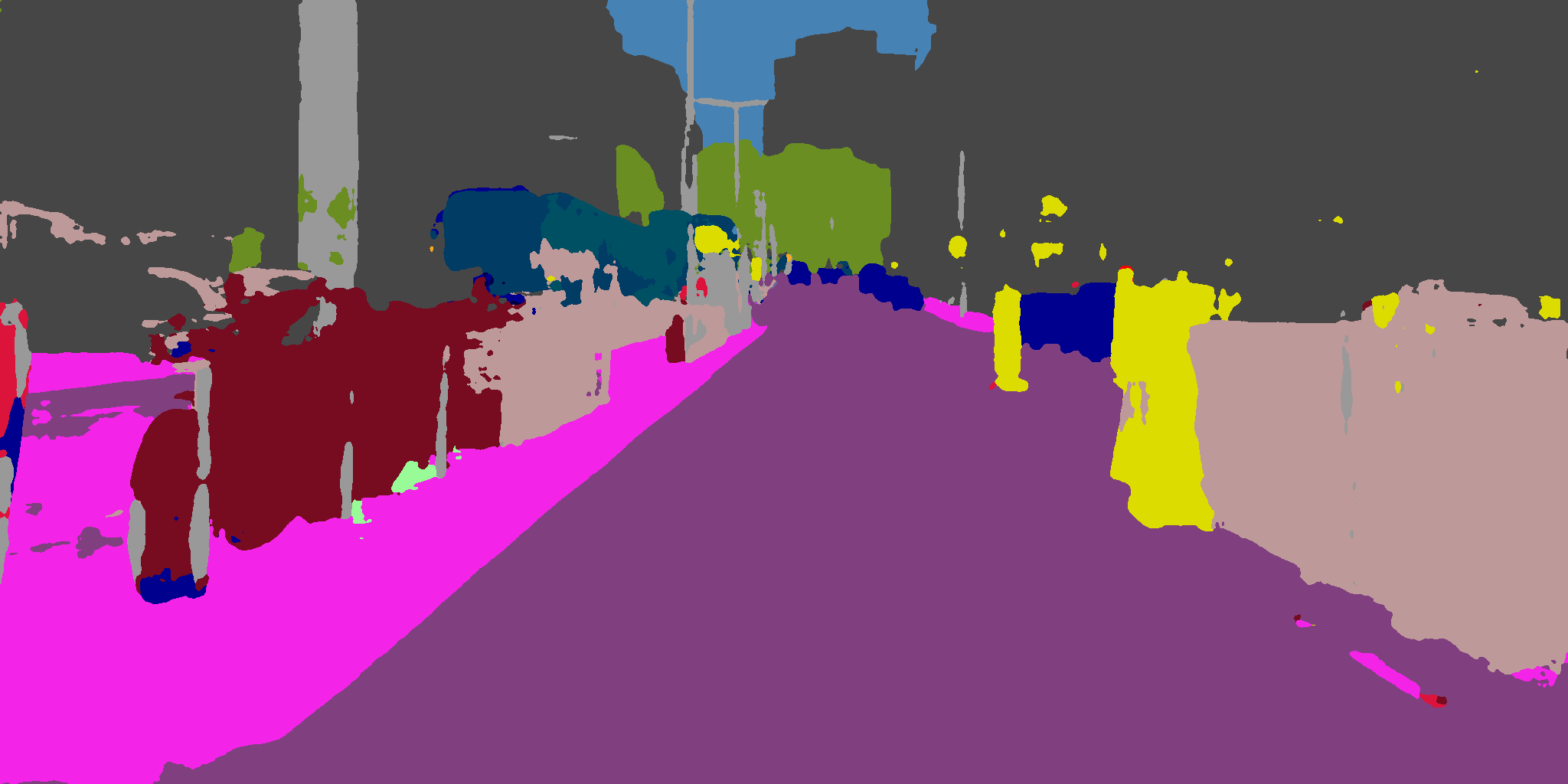}} &
			\subfloat{\includegraphics[width = 0.19\linewidth]{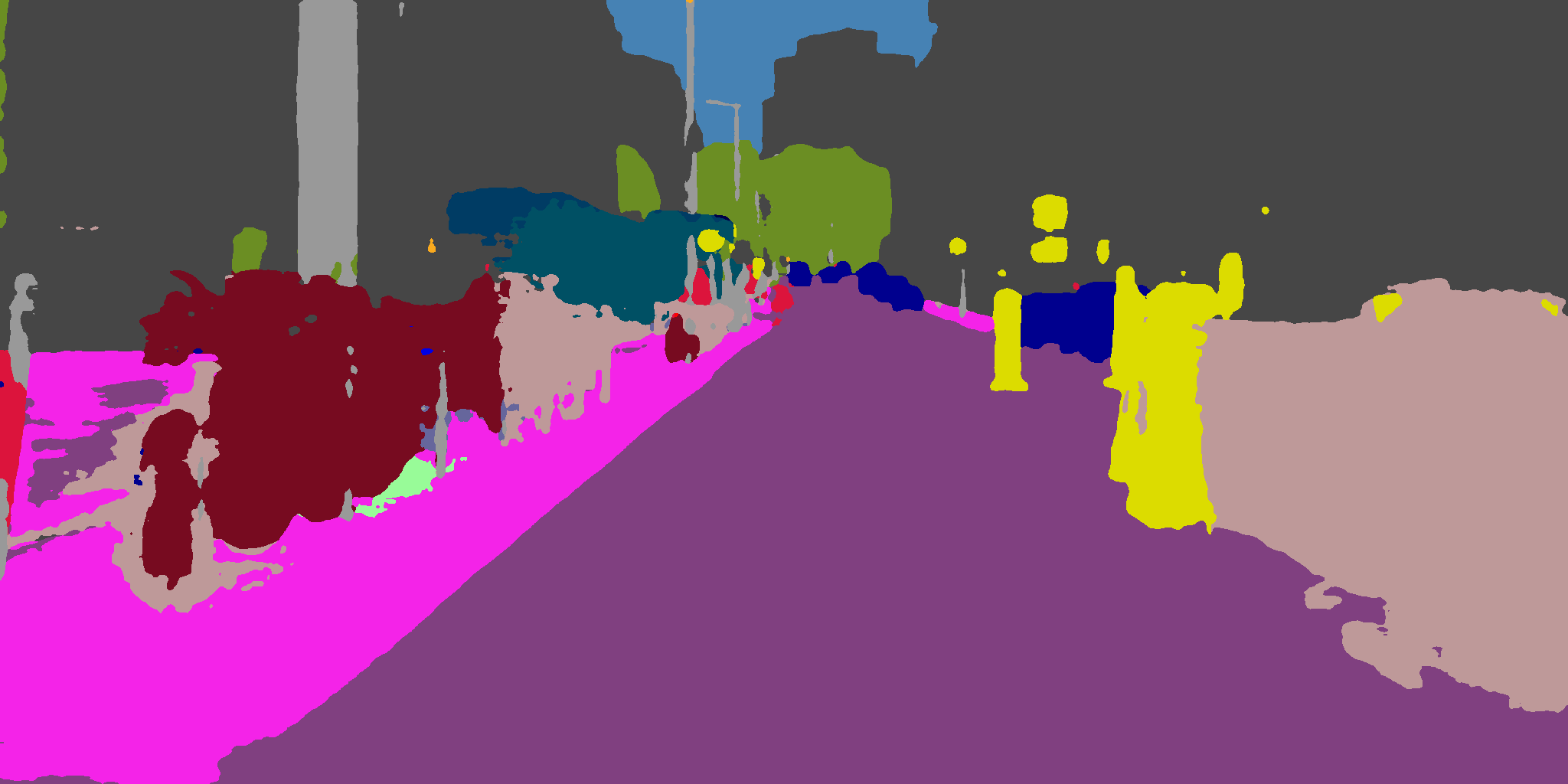}}\\
			Image&GT&arch0&arch1
	\end{tabular}}
	\caption{Qualitative results of the discovered models - (\emph{arch0} and \emph{arch1}) - on the validation set of CityScapes. The last row shows failure cases.}
	\label{fig:cs-res}
\end{figure*}

\section{Training Experiments}
\label{sec:training}
\subsection{Setup}

To evaluate the best discovered architectures, we consider two common benchmarks for semantic segmentation -- CityScapes~\cite{CordtsORREBFRS16} and CamVid~\cite{BrostowFC09}. Our choice is motivated by the fact that the majority of hand-designed compact networks is being extensively tested on these two datasets.

The training setup slightly differs from the searching one: in particular, we use the `poly' training schedule~\cite{ChenPKMY18} -- $lr_{init}\cdot(1 - \frac{epoch}{n_{epochs}})^{0.9}$ -- with SGD with the initial learning rate of $5{e-}2$ and the momentum value of $0.9$. The weight decay is set to $1{e-}5$, the initial number of channels to $64$.

\subsection{CityScapes}
We train on the full training split and test on the validation split of $500$ images. Here we use a square crop size of $769$ and train for $1000$ epochs with mini-batches of $6$ examples.

As given in Table~\ref{table:cs}, two of our automatically discovered models achieve $67.7\%$ and $67.8\%$ mean IoU, respectively, with both having less than $300$K trainable parameters. We outperform all other compact methods with the exclusion of ERFNet~\cite{RomeraABA18} that comprises $7\times$ more parameters than any of our models, and BiSeNet~\cite{YuWPGYS18} - with more than $20{\times}$ more parameters. We overcome ICNet~\cite{ZhaoQSSJ18} on the validation set, but fall behind on the test set as their method was further trained on the validation set, too. Furthermore, we significantly surpass the results of ESPNet~\cite{MehtaRCSH18} and ESPNet{-}v2~\cite{abs-1811-11431} by more than $5.4\%$ in both cases while requiring considerably fewer parameters.

\begin{table}[htb]
	\begin{center}
		\begin{adjustbox}{max width=0.45\textwidth}
			\begin{tabular}{l|c|c|c}
				\specialrule{.15em}{0em}{0em}
				Method & val mIoU,\% & test mIoU,\% & Params,M\T\B\\
				\specialrule{.1em}{0em}{0em}
				ENet~\cite{PaszkeCKC16} & - & 58.3 & 0.37\T\B\\
				ESPNet~\cite{MehtaRCSH18} & 61.4 & 60.3 & 0.36\T\B\\
				ESPNet{-}v2~\cite{abs-1811-11431} & 62.7 & 62.1 &  0.72\T\B\\
				ICNet~\cite{ZhaoQSSJ18} & 67.7 & 69.5 & 6.7\T\B\\
				ERFNet~\cite{RomeraABA18} & 71.5 & 69.7 & 2.1\T\B\\
				BiSeNet~\cite{YuWPGYS18} & \textbf{72.0} & \textbf{71.4} &  5.8\B\\
				\hline
				Ours (arch0) & 68.1 & 67.7\tablefootnote{Link to test results: \url{https://bit.ly/2HItlwm}} & \textbf{0.28}\T\B\\
				Ours (arch1) & 69.5 & 67.8\tablefootnote{Link to test results: \url{https://bit.ly/2FlAfEW}} & \textbf{0.27}\T\B\\
				\specialrule{.15em}{0em}{0em}
			\end{tabular}
		\end{adjustbox}
		\caption{Quantitative results on the validation and test sets of CityScapes among compact models (${<}10$M parameters). Note that opposed to what is commonly done, we did not train our models on the val set and did not use any post-processing for test evaluation. %
			\label{table:cs}}
	\end{center}
	\vskip -0.05in
\end{table}

We visualise qualitative results in Fig.~\ref{fig:cs-res}. Both architectures are able to correctly segment most parts of the scenes and even identify thin structures such as traffic lights and poles (rows $1{-}2$). Nevertheless, they tend to misclassify large objects, such as trains (row $3$).

\subsection{CamVid}
CamVid~\cite{BrostowFC09} is another outdoor urban driving dataset that contains $367$ images for training and $233$ -- for testing with $11$ semantic classes and resolution of $480{\times}360$. We train for $1000$ epochs with $10$ examples in mini-batch.

The architectures discovered by our method attain mean IoU values of $63.9\%$ and $63.2\%$, respectively, once again exceeding the majority of other compact and even larger models. We outperform both SegNet~\cite{BadrinarayananK17} and ESPNet~\cite{MehtaRCSH18} by more than $7\%$, DeepLab-LFOV~\cite{ChenPKMY18} -- by more than $1.6\%$, and fall behind BiSeNet~\cite{YuWPGYS18} and ICNet~\cite{ZhaoQSSJ18}, both of which exploited higher resolution images of $960{\times}720$.

\begin{table}[htb]
	\begin{center}
		\begin{adjustbox}{max width=0.45\textwidth}
			\begin{tabular}{l|c|c}
				\specialrule{.15em}{0em}{0em}
				Method & mIoU,\% & Params,M\T\B\\
				\specialrule{.1em}{0em}{0em}
				SegNet~\cite{BadrinarayananK17} & 55.6 & 29.7\T\B\\
				ESPNet~\cite{MehtaRCSH18} & 55.6 & 0.36\T\B\\
				DeepLab-LFOV~\cite{ChenPKMY18} & 61.6 & 37.3\T\B\\
				$\dagger$BiSeNet~\cite{YuWPGYS18} & \textbf{65.6} & 5.8\T\B\\
				$\dagger$ICNet~\cite{ZhaoQSSJ18} & \textbf{67.1} & 6.7\T\B\\
				\hline
				Ours (arch0) & 63.9 & \textbf{0.28}\T\B\\
				Ours (arch1) & 63.2 & \textbf{0.26}\T\B\\
				\specialrule{.15em}{0em}{0em}
			\end{tabular}
		\end{adjustbox}
		\caption{Quantitative results on the test set of CamVid. ($\dagger$) means that $960{\times}720$ images were used opposed to $480{\times}360$. %
			\label{table:cv}}
	\end{center}
	\vskip -0.15in
\end{table}

\begin{figure*}[t]
	\centering
	\resizebox{1.\textwidth}{!}{\begin{tabular}{cc|cc}
			\subfloat{\includegraphics[width = 0.19\linewidth]{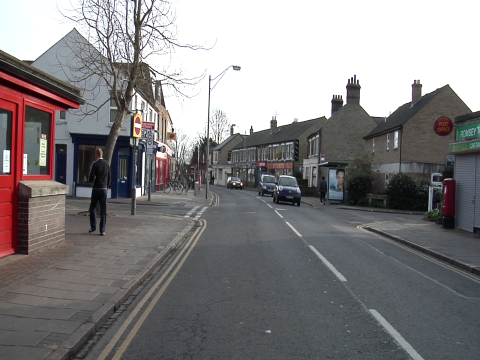}} &
			\subfloat{\includegraphics[width = 0.19\linewidth]{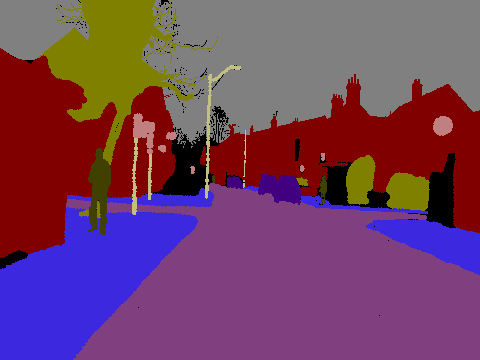}} &
			\subfloat{\includegraphics[width = 0.19\linewidth]{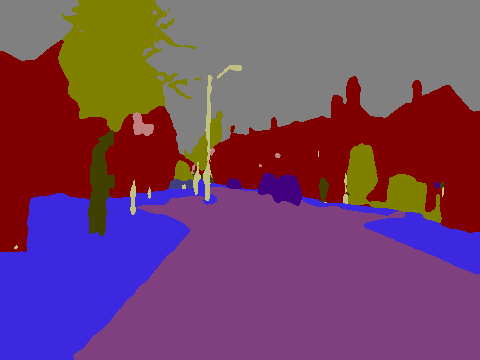}} &
			\subfloat{\includegraphics[width = 0.19\linewidth]{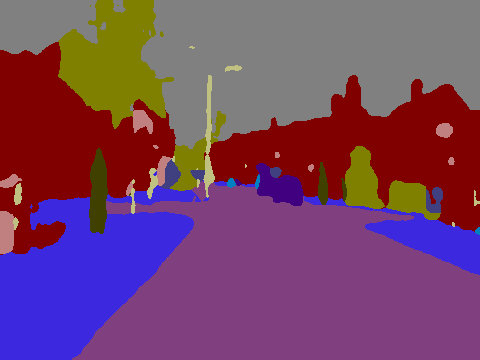}}\\[-0.15in]
			\subfloat{\includegraphics[width = 0.19\linewidth]{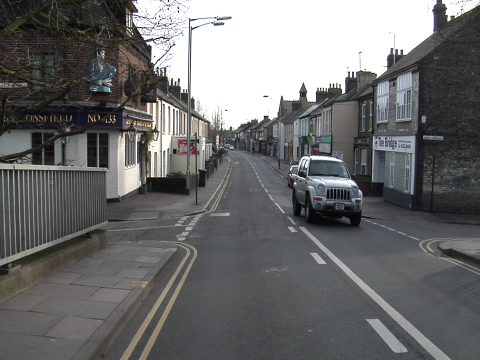}} &
			\subfloat{\includegraphics[width = 0.19\linewidth]{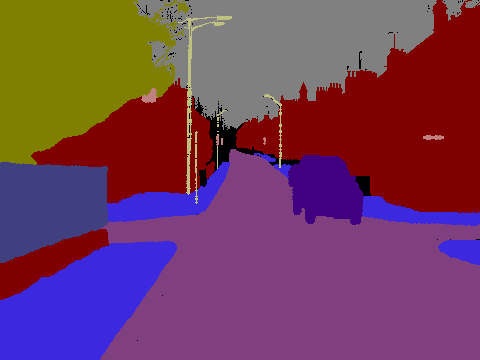}} &
			\subfloat{\includegraphics[width = 0.19\linewidth]{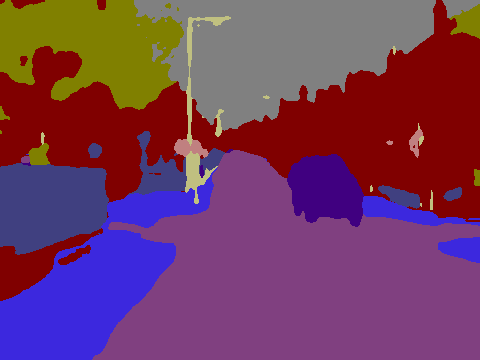}} &
			\subfloat{\includegraphics[width = 0.19\linewidth]{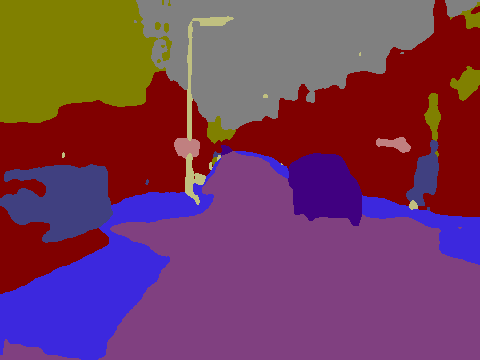}}\\[-0.15in]
			\subfloat{\includegraphics[width = 0.19\linewidth]{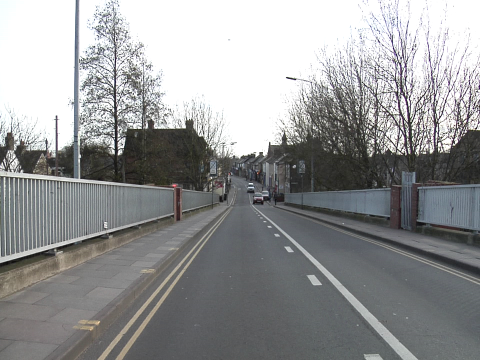}} &
			\subfloat{\includegraphics[width = 0.19\linewidth]{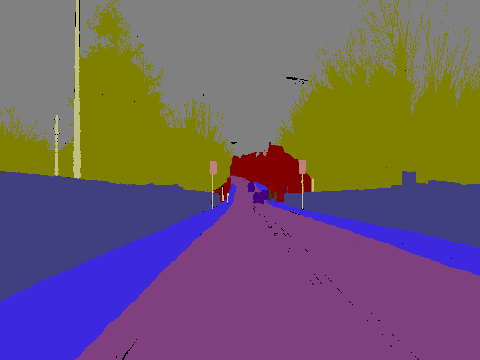}} &
			\subfloat{\includegraphics[width = 0.19\linewidth]{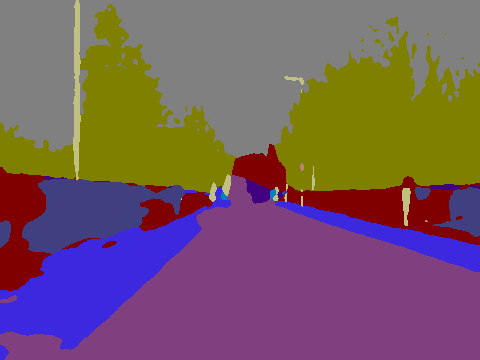}} &
			\subfloat{\includegraphics[width = 0.19\linewidth]{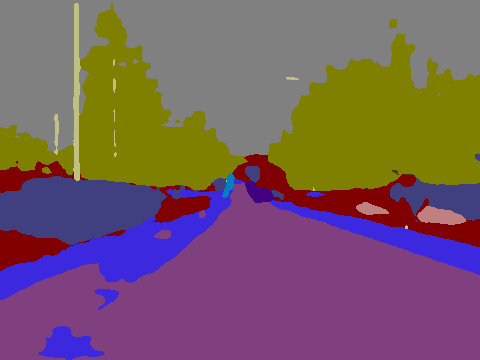}}\\    
			Image&GT&arch0&arch1
	\end{tabular}}
	\caption{Qualitative results of the discovered models - (\emph{arch0} and \emph{arch1}) - on the test set of CamVid. Last row includes failure cases.}
	\label{fig:cv-res}
	\vskip -0.1in
\end{figure*}

With the lower quality and resolution of annotations, the predictions are no longer sharp as on CityScapes (Fig.~\ref{fig:cv-res}). Overall, both architectures capture the semantics of the scenes well (rows $1{-}2$) and only fail significantly at delineating long chunks of the same class on the edges of the image (row $3$).

\subsection{Architecture Characteristics}

We provide quantitative details of the trained architectures in Table~\ref{table:char}. Two models possess similar qualities -- they are light-weight, both in terms of parameters and the size on disk, and have output resolution of $\frac{1}{4}$. Interestingly, even though {\em arch0} is slightly larger than {\em arch1}, it is still faster with almost $20$FPS on high-resolution $2048{\times}1024$ images.\footnote{In the first version of the paper we wrongly claimed that the runtime measurements of arch0 and arch1 were 95.7$\pm$0.721 and 145$\pm$0.215, correspondingly. Those numbers were incorrect as the model was in the training regime and not in the evaluation mode of PyTorch.} Below are the characteristics of the discovered architectures with correct latency: This indicates that the connectivity structure plays an important role in determining the runtime of a network, which may signal that to enforce a real-time constraint, one would need to include an explicit loss term in the RL objective -- e.g. as done in~\cite{WuDZWSWTVJK19,TanCPVSHL19}.
\begin{table}[htb]
	\begin{center}
		\begin{adjustbox}{max width=0.45\textwidth}
			\begin{tabular}{c|c|c}
				\specialrule{.15em}{0em}{0em}
				Characteristic & arch0 & arch1 \T\B\\
				\specialrule{.1em}{0em}{0em}
				Parameters,K & 280.15 & \textbf{268.24}\T\B\\
				Latency, ms & \textbf{52.25$\pm$0.003} & 97.11$\pm$0.24\T\B\\
				Output Resolution & $1/4$ & $1/4$ \T\B\\
				Size on disk, MB & 1.46 & \textbf{1.41}\B\\
				\specialrule{.15em}{0em}{0em}
			\end{tabular}
		\end{adjustbox}
		\caption{Quantitative characteristics of discovered architectures. All the numbers are measured on $2048{\times}1024$ resolution with $19$ output classes using a single $1080$Ti GPU.
			\label{table:char}}
	\end{center}
	\vskip -0.15in
\end{table}

We further visualise one of discovered architectures -- {\em arch0}\footnote{The illustration of {\em arch1} is provided in Appendix.} -- on Fig.~\ref{fig:arch0}. Notably, all the templates rely on concatenation with the generated structure having multiple connections between intermediate blocks that allows the network to simultaneously operate on high-level and low-level features.

\begin{figure}[thb]
	\begin{center}
		\includegraphics[width=1.\linewidth]{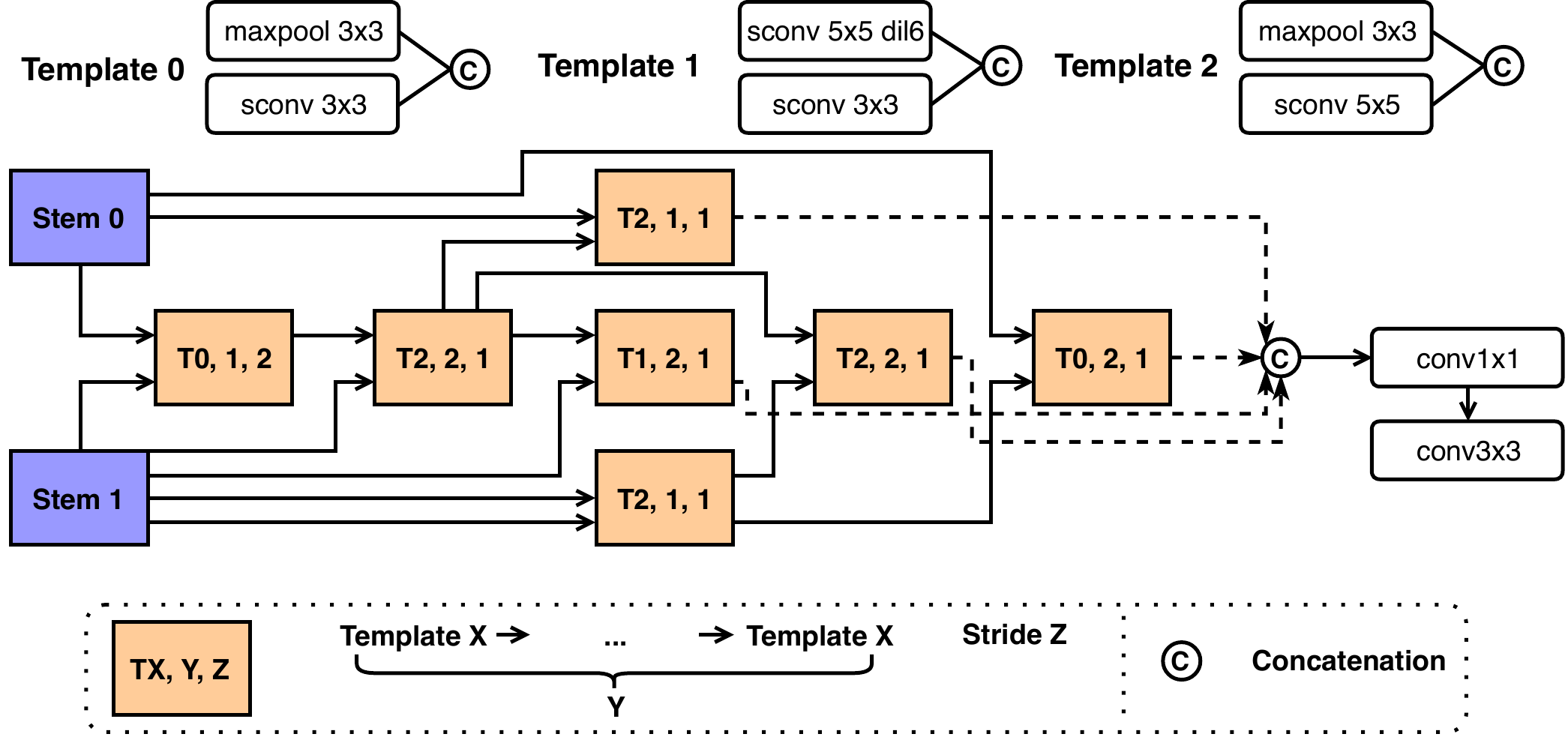}
	\end{center}
	\vskip -0.1in
	\caption{Depiction of {\em arch0}.\label{fig:arch0}}
	\vskip -0.3in
\end{figure}

\section{Conclusions}
\vskip -0.085in
In this work we tackled the problem of automatic architecture search of light-weight segmentation networks. Previous approaches carried one significant disadvantage of being extremely reliant on complete image classification networks. Here, we overcome this issue by proposing a solution that only requires a small portion of the image classifier, with roughly $60$K parameters. To this end, our approach decomposes the sequence of decisions to make into templates which are sets of operations that can be applied anywhere in the network. Besides, we also predict how many times the templates should be used and what downsampling factor they should use.

In an extensive set of experiments we showcased that our search process reliably predicts promising and high-performing architectures. The full training experiments further confirmed the search results as the generated models achieved $63.9\%$ and $67.8\%$ mean IoU on CamVid and CityScapes, respectively, while containing less than $300$K parameters. As a result, we significantly outperformed other manually-designed compact solutions. In future work, we will concentrate on removing the need for a pre-trained classifier altogether and adding an explicit real-time constraint, which may lead to even better solutions.\\

\noindent\textbf{Acknowledgements}
    VN, CS, IR's participation in this work were in part supported by ARC Centre of Excellence for Robotic Vision.

\clearpage
{\small
	\bibliographystyle{ieee}
	\bibliography{egbib}

\begin{thebibliography}{10}\itemsep=-1pt

\bibitem{BadrinarayananK17}
V.~Badrinarayanan, A.~Kendall, and R.~Cipolla.
\newblock Segnet: {A} deep convolutional encoder-decoder architecture for image
  segmentation.
\newblock {\em {IEEE} Trans. Pattern Anal. Mach. Intell.}, 39(12):2481--2495,
  2017.

\bibitem{BrostowFC09}
G.~J. Brostow, J.~Fauqueur, and R.~Cipolla.
\newblock Semantic object classes in video: {A} high-definition ground truth
  database.
\newblock {\em Pattern Recognition Letters}, 30(2):88--97, 2009.

\bibitem{abs-1809-04184}
L.~Chen, M.~D. Collins, Y.~Zhu, G.~Papandreou, B.~Zoph, F.~Schroff, H.~Adam,
  and J.~Shlens.
\newblock Searching for efficient multi-scale architectures for dense image
  prediction.
\newblock {\em Proc. Advances in Neural Inf. Process. Syst.}, 2018.

\bibitem{ChenPKMY18}
L.~Chen, G.~Papandreou, I.~Kokkinos, K.~Murphy, and A.~L. Yuille.
\newblock Deeplab: Semantic image segmentation with deep convolutional nets,
  atrous convolution, and fully connected crfs.
\newblock {\em {IEEE} Trans. Pattern Anal. Mach. Intell.}, 2018.

\bibitem{ChenZPSA18}
L.~Chen, Y.~Zhu, G.~Papandreou, F.~Schroff, and H.~Adam.
\newblock Encoder-decoder with atrous separable convolution for semantic image
  segmentation.
\newblock In {\em Proc. Eur. Conf. Comp. Vis.}, 2018.

\bibitem{CordtsORREBFRS16}
M.~Cordts, M.~Omran, S.~Ramos, T.~Rehfeld, M.~Enzweiler, R.~Benenson,
  U.~Franke, S.~Roth, and B.~Schiele.
\newblock The cityscapes dataset for semantic urban scene understanding.
\newblock In {\em Proc. IEEE Conf. Comp. Vis. Patt. Recogn.}, 2016.

\bibitem{HeZRS16}
K.~He, X.~Zhang, S.~Ren, and J.~Sun.
\newblock Deep residual learning for image recognition.
\newblock In {\em Proc. IEEE Conf. Comp. Vis. Patt. Recogn.}, 2016.

\bibitem{KandasamyNSPX18}
K.~Kandasamy, W.~Neiswanger, J.~Schneider, B.~P{\'{o}}czos, and E.~P. Xing.
\newblock Neural architecture search with bayesian optimisation and optimal
  transport.
\newblock In {\em Proc. Advances in Neural Inf. Process. Syst.}, 2018.

\bibitem{LinMSR17}
G.~Lin, A.~Milan, C.~Shen, and I.~D. Reid.
\newblock {RefineNet}: Multi-path refinement networks for high-resolution
  semantic segmentation.
\newblock In {\em Proc. IEEE Conf. Comp. Vis. Patt. Recogn.}, 2017.

\bibitem{abs-1901-02985}
C.~Liu, L.~Chen, F.~Schroff, H.~Adam, W.~Hua, A.~L. Yuille, and L.~Fei{-}Fei.
\newblock Auto-deeplab: Hierarchical neural architecture search for semantic
  image segmentation.
\newblock {\em arXiv: Comp. Res. Repository}, abs/1901.02985, 2019.

\bibitem{abs-1711-00436}
H.~Liu, K.~Simonyan, O.~Vinyals, C.~Fernando, and K.~Kavukcuoglu.
\newblock Hierarchical representations for efficient architecture search.
\newblock {\em Proc. Int. Conf. Learn. Representations}, 2018.

\bibitem{abs-1806-09055}
H.~Liu, K.~Simonyan, and Y.~Yang.
\newblock {DARTS:} differentiable architecture search.
\newblock {\em arXiv: Comp. Res. Repository}, abs/1806.09055, 2018.

\bibitem{LongSD15}
J.~Long, E.~Shelhamer, and T.~Darrell.
\newblock Fully convolutional networks for semantic segmentation.
\newblock In {\em Proc. IEEE Conf. Comp. Vis. Patt. Recogn.}, 2015.

\bibitem{MehtaRCSH18}
S.~Mehta, M.~Rastegari, A.~Caspi, L.~G. Shapiro, and H.~Hajishirzi.
\newblock Espnet: Efficient spatial pyramid of dilated convolutions for
  semantic segmentation.
\newblock In {\em Proc. Eur. Conf. Comp. Vis.}, 2018.

\bibitem{abs-1811-11431}
S.~Mehta, M.~Rastegari, L.~G. Shapiro, and H.~Hajishirzi.
\newblock Espnetv2: {A} light-weight, power efficient, and general purpose
  convolutional neural network.
\newblock {\em arXiv: Comp. Res. Repository}, abs/1811.11431, 2018.

\bibitem{abs-1810-10804}
V.~Nekrasov, H.~Chen, C.~Shen, and I.~D. Reid.
\newblock Fast neural architecture search of compact semantic segmentation
  models via auxiliary cells.
\newblock {\em Proc. IEEE Conf. Comp. Vis. Patt. Recogn.}, 2019.

\bibitem{PaszkeCKC16}
A.~Paszke, A.~Chaurasia, S.~Kim, and E.~Culurciello.
\newblock Enet: {A} deep neural network architecture for real-time semantic
  segmentation.
\newblock {\em arXiv: Comp. Res. Repository}, abs/1606.02147, 2016.

\bibitem{PhamGZLD18}
H.~Pham, M.~Y. Guan, B.~Zoph, Q.~V. Le, and J.~Dean.
\newblock Efficient neural architecture search via parameter sharing.
\newblock In {\em Proc. Int. Conf. Mach. Learn.}, 2018.

\bibitem{PohlenHML17}
T.~Pohlen, A.~Hermans, M.~Mathias, and B.~Leibe.
\newblock Full-resolution residual networks for semantic segmentation in street
  scenes.
\newblock In {\em Proc. IEEE Conf. Comp. Vis. Patt. Recogn.}, 2017.

\bibitem{RealMSSSTLK17}
E.~Real, S.~Moore, A.~Selle, S.~Saxena, Y.~L. Suematsu, J.~Tan, Q.~V. Le, and
  A.~Kurakin.
\newblock Large-scale evolution of image classifiers.
\newblock In {\em Proc. Int. Conf. Mach. Learn.}, 2017.

\bibitem{RomeraABA18}
E.~Romera, J.~M. Alvarez, L.~M. Bergasa, and R.~Arroyo.
\newblock Erfnet: Efficient residual factorized convnet for real-time semantic
  segmentation.
\newblock {\em {IEEE} Trans. Intelligent Transportation Systems}, 19(1), 2018.

\bibitem{RonnebergerFB15}
O.~Ronneberger, P.~Fischer, and T.~Brox.
\newblock U-net: Convolutional networks for biomedical image segmentation.
\newblock In {\em MICCAI}, 2015.

\bibitem{RussakovskyDSKS15}
O.~Russakovsky, J.~Deng, H.~Su, J.~Krause, S.~Satheesh, S.~Ma, Z.~Huang,
  A.~Karpathy, A.~Khosla, M.~S. Bernstein, A.~C. Berg, and F.~Li.
\newblock Imagenet large scale visual recognition challenge.
\newblock {\em Int. J. Comput. Vision}, 115(3), 2015.

\bibitem{abs-1801-04381}
M.~Sandler, A.~G. Howard, M.~Zhu, A.~Zhmoginov, and L.~Chen.
\newblock Inverted residuals and linear bottlenecks: Mobile networks for
  classification, detection and segmentation.
\newblock {\em Proc. IEEE Conf. Comp. Vis. Patt. Recogn.}, 2018.

\bibitem{SchulmanWDRK17}
J.~Schulman, F.~Wolski, P.~Dhariwal, A.~Radford, and O.~Klimov.
\newblock Proximal policy optimization algorithms.
\newblock {\em arXiv: Comp. Res. Repository}, abs/1707.06347, 2017.

\bibitem{TanCPVSHL19}
M.~Tan, B.~Chen, R.~Pang, V.~Vasudevan, M.~Sandler, A.~Howard, and Q.~V. Le.
\newblock Mnasnet: Platform-aware neural architecture search for mobile.
\newblock In {\em Proc. IEEE Conf. Comp. Vis. Patt. Recogn.}, 2019.

\bibitem{WuDZWSWTVJK19}
B.~Wu, X.~Dai, P.~Zhang, Y.~Wang, F.~Sun, Y.~Wu, Y.~Tian, P.~Vajda, Y.~Jia, and
  K.~Keutzer.
\newblock Fbnet: Hardware-aware efficient convnet design via differentiable
  neural architecture search.
\newblock In {\em Proc. IEEE Conf. Comp. Vis. Patt. Recogn.}, 2019.

\bibitem{YuWPGYS18}
C.~Yu, J.~Wang, C.~Peng, C.~Gao, G.~Yu, and N.~Sang.
\newblock Bisenet: Bilateral segmentation network for real-time semantic
  segmentation.
\newblock In {\em Proc. Eur. Conf. Comp. Vis.}, 2018.

\bibitem{corr/YuK15}
F.~Yu and V.~Koltun.
\newblock Multi-scale context aggregation by dilated convolutions.
\newblock {\em Proc. Int. Conf. Learn. Representations}, 2016.

\bibitem{ZhaoQSSJ18}
H.~Zhao, X.~Qi, X.~Shen, J.~Shi, and J.~Jia.
\newblock Icnet for real-time semantic segmentation on high-resolution images.
\newblock In {\em Proc. Eur. Conf. Comp. Vis.}, 2018.

\bibitem{ZhaoSQWJ17}
H.~Zhao, J.~Shi, X.~Qi, X.~Wang, and J.~Jia.
\newblock Pyramid scene parsing network.
\newblock In {\em Proc. IEEE Conf. Comp. Vis. Patt. Recogn.}, 2017.

\bibitem{ZophL16}
B.~Zoph and Q.~V. Le.
\newblock Neural architecture search with reinforcement learning.
\newblock {\em Proc. Int. Conf. Learn. Representations}, 2017.

\bibitem{ZophVSL17}
B.~Zoph, V.~Vasudevan, J.~Shlens, and Q.~V. Le.
\newblock Learning transferable architectures for scalable image recognition.
\newblock {\em Proc. IEEE Conf. Comp. Vis. Patt. Recogn.}, 2018.

\end{thebibliography}
}

 \clearpage
 
 \section*{Supplementary Material}
 
 \section{Analysis of Search Results}
 In addition to the discussion in the main text, we are following up on two more questions: i.) whether there appears to be any correlation between the number of parameters of a sampled architecture and its performance, and ii.) what templates lead to larger rewards.

 \subsection*{Number of Parameters}
 
\begin{figure}[thb]
 	\begin{center}
 		\includegraphics[width=1.\linewidth]{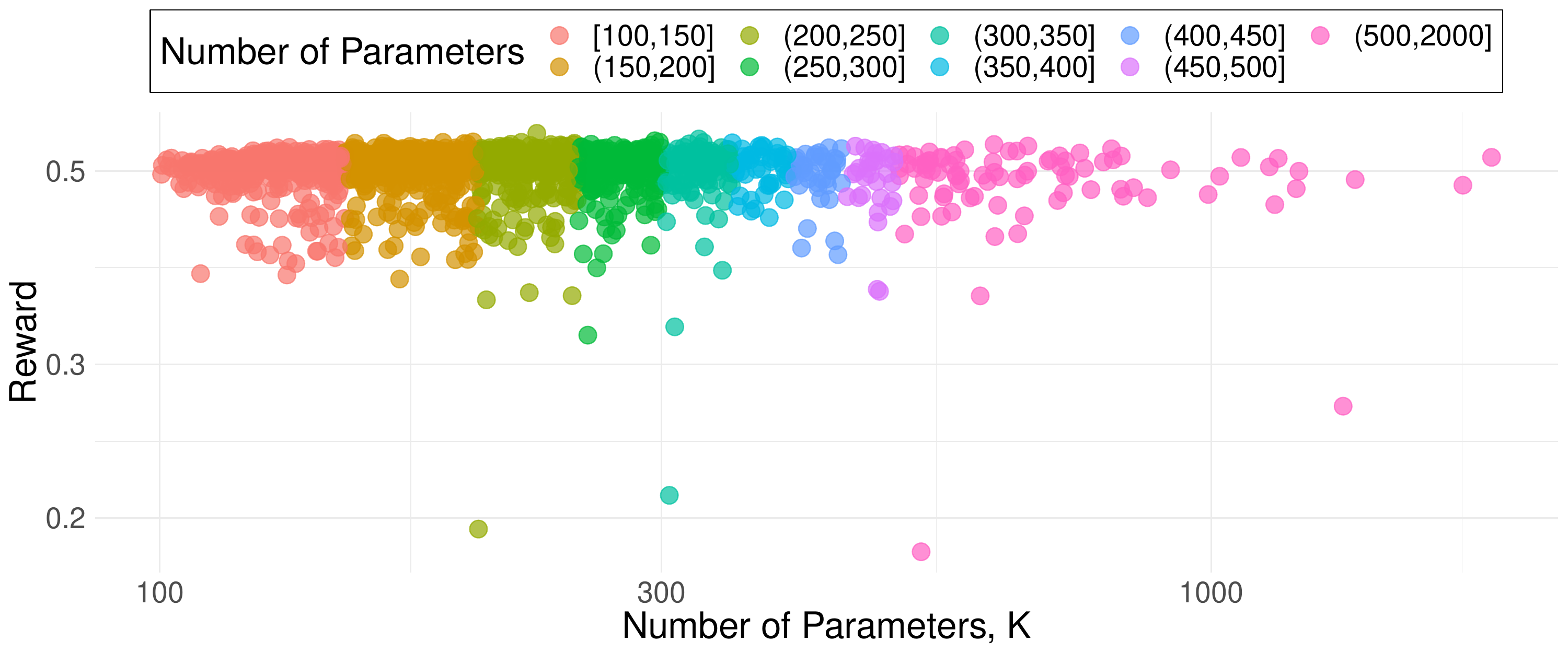}
 	\end{center}
 	\caption{Reward as a function of the size of the architectures.\label{fig:rew-all}}
 \end{figure}

\begin{figure}[thb]
 	\begin{center}
 		\includegraphics[width=1.\linewidth]{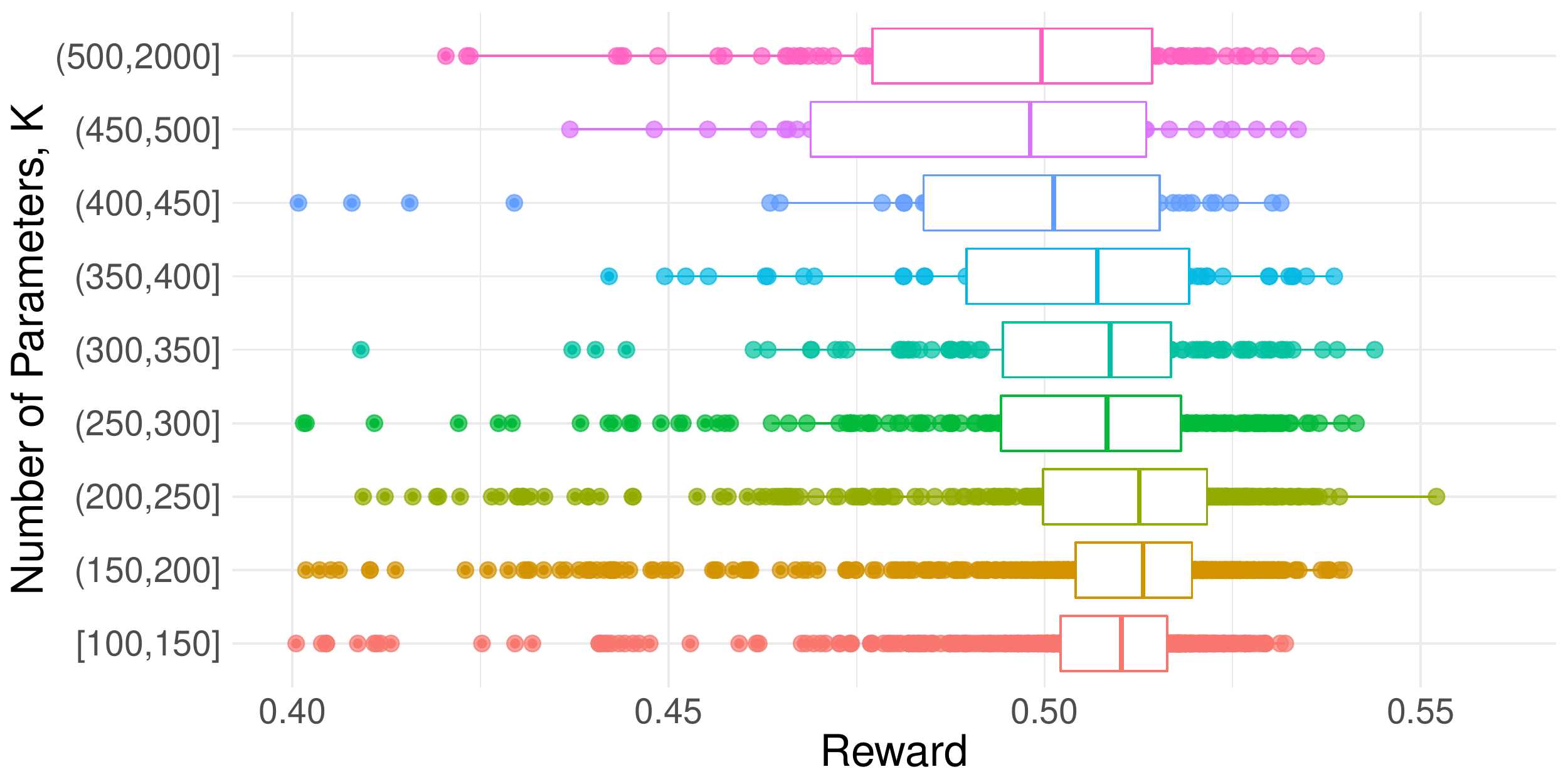}
 	\end{center}
 	\caption{
 		Distribution of rewards attained by architectures with varying size. For compactness of the plot, we only visualise rewards greater than or equal to $0.40$.\label{fig:rew-grouped}}
 \end{figure}

 First, we consider the distribution of rewards based on the number of parameters (Fig.~\ref{fig:rew-all}). From it, the size of the architecture appears to have no connection with its reward. A more detailed plot tells a different story, though (Fig.~\ref{fig:rew-grouped}): while for small architectures ($\leq{250}$K) the rewards are almost identically distributed, a negative trend can be seen when the number of parameters is growing. It is possible that this effect occurs due to the utilised training strategy favouring compact architectures. 

 \subsection*{Templates}
 
 We remind the reader that during the search process the controller samples $3$ template structures that can be applied at any of $7$ blocks $1{-}4$ times recursively. Each template consists of two individual operations and one aggregation operation. In total, there are $6$ unique operations and $2$ unique aggregation operations, which leads to $C_{6+1}^{2}=\frac{7!}{5!2!}=42$ unique templates taking into account the symmetry in the order of individual operations.
 
\begin{figure}[thb]
 	\begin{center}
 		\includegraphics[width=1.\linewidth]{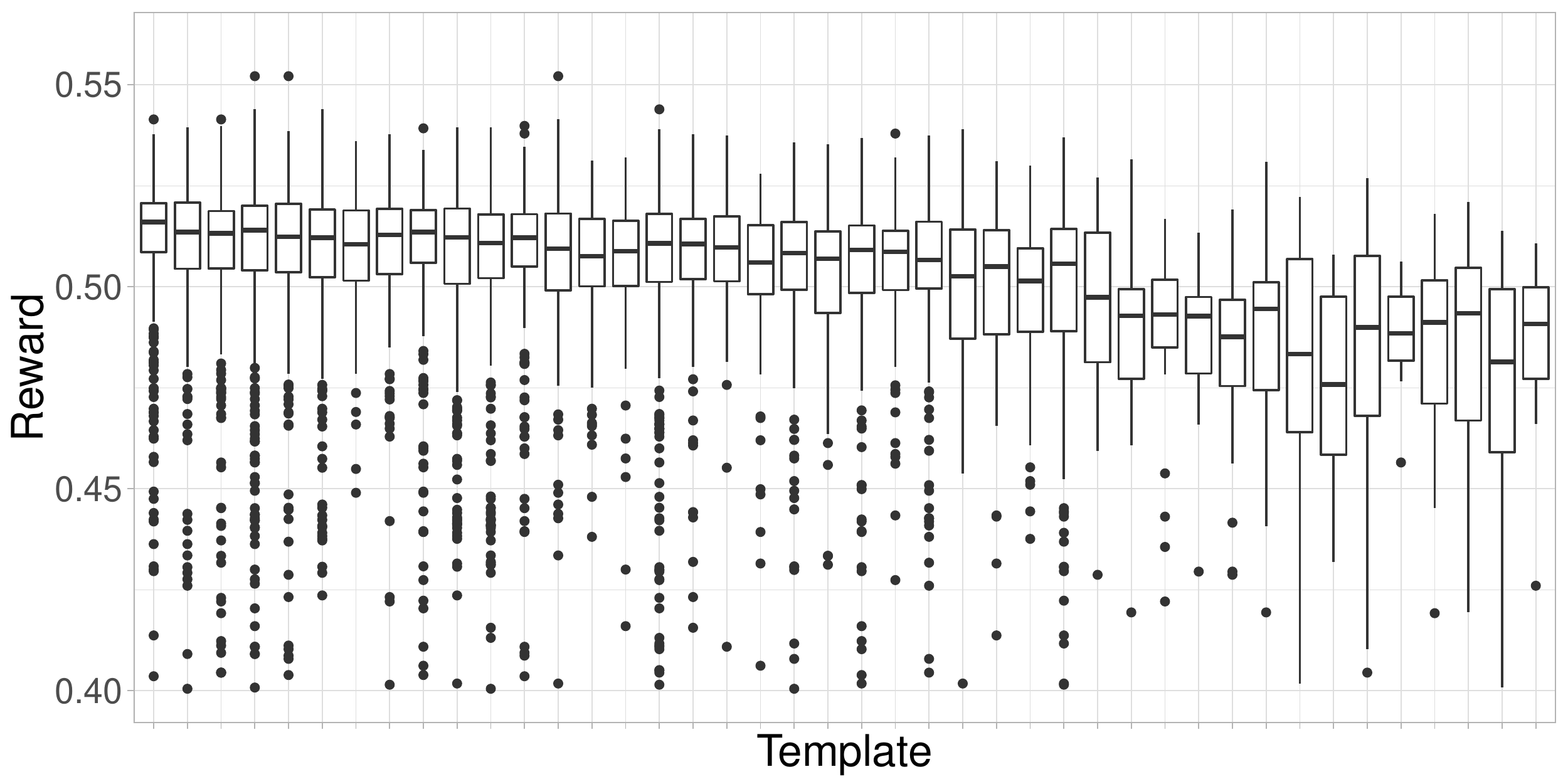}
 	\end{center}
 	\vskip -0.2in
 	\caption{
 		Distribution of rewards for each of $42$ unique templates. For compactness of the plot, we only visualise rewards greater than or equal to $0.40$.\label{fig:temp-rew-all}}
 \end{figure}

 We consider a particular template to be sampled during the search process, if its structure was sampled and it was chosen by the controller at least once. We visualise the distribution of rewards for each of $42$ templates sampled during the search process in Fig.~\ref{fig:temp-rew-all} - note that several templates can share the same reward as they might belong to a single architecture. Overall, around half of all the templates steadily achieve rewards higher than $0.5$.
 
\begin{figure}[thb]
 	\begin{center}
 		\includegraphics[width=1.\linewidth]{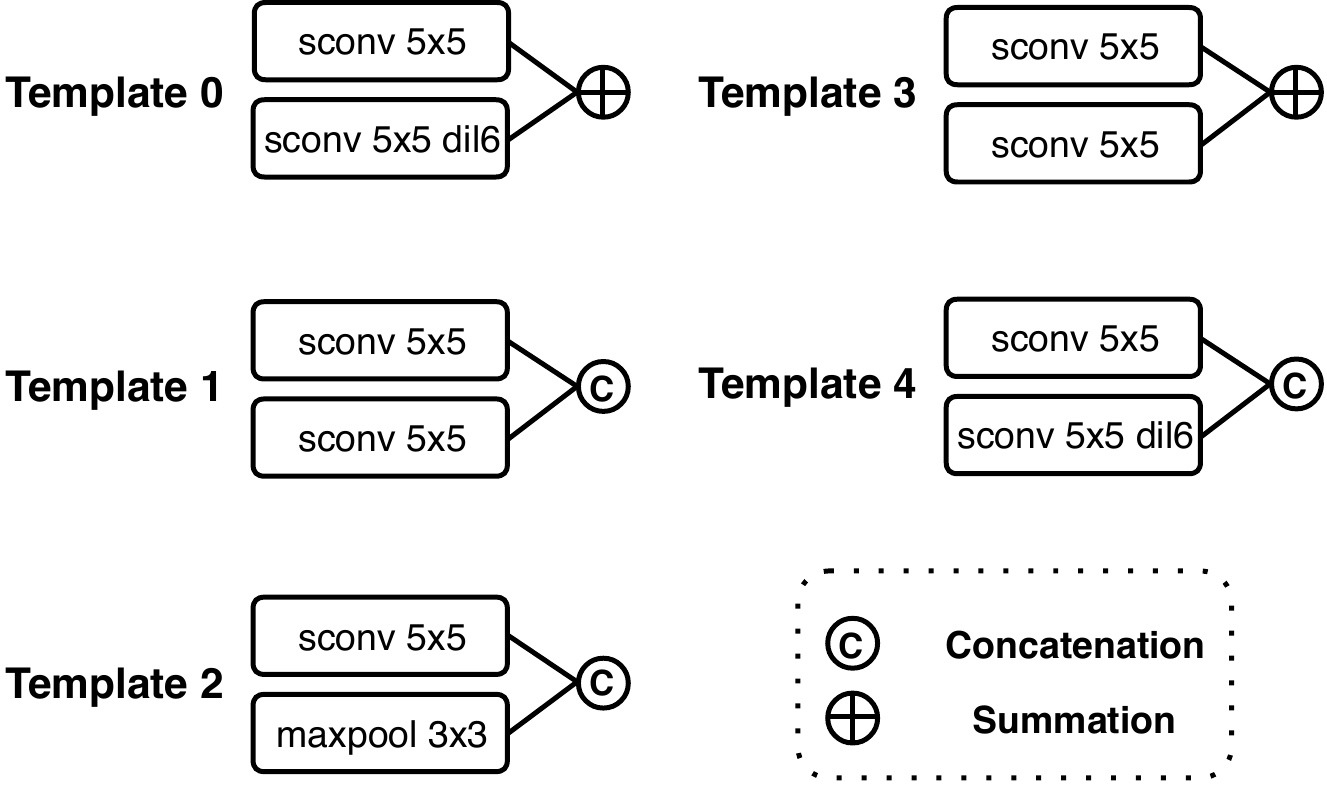}
 	\end{center}
 	\caption{
 		Top-$5$ templates with highest average reward.\label{fig:temp-top}}
 \end{figure}

 We further depict each of top-$5$ templates with highest rewards in Fig.~\ref{fig:temp-top}. Interestingly, all top-performing templates rely on separable $5{\times}5$ convolution, and some of them differ only in the aggregation operation used (\eg~{\em Template 0} and {\em Template 4}, or {\em Template 1} and {\em Template 3}). 
 
 \section{Architecture Characteristics}
\begin{figure}[thb]
 	\begin{center}
 		\includegraphics[width=1.\linewidth]{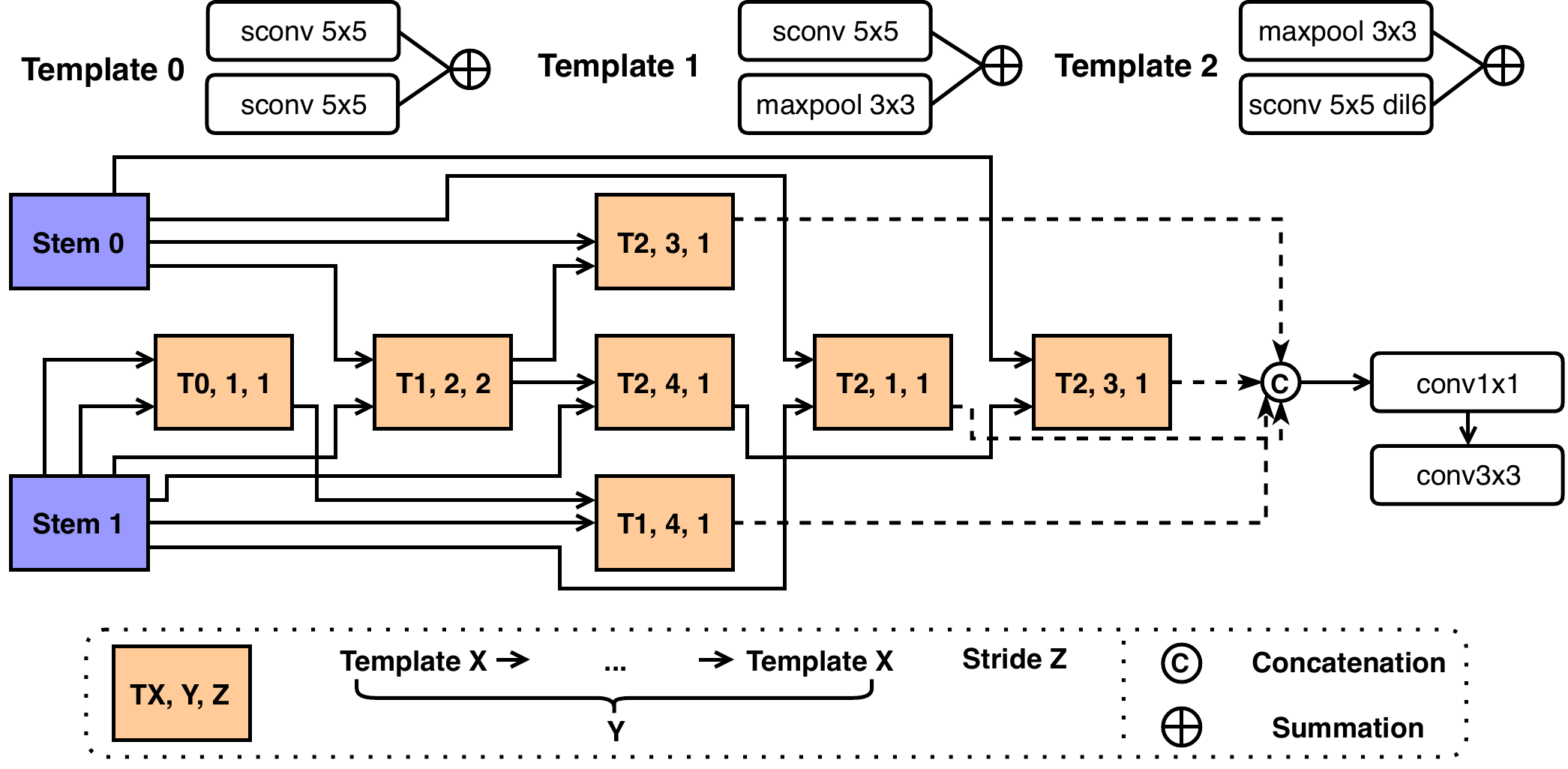}
 	\end{center}
 	\caption{Depiction of {\em arch1}.\label{fig:arch1}}
 \end{figure}

 We visualise another of discovered architectures - {\em arch1} - in Fig.~\ref{fig:arch1} (please refer to the main text for the visualisation of {\em arch0}). In contrast to {\em arch0}, this architecture fully relies on the summation as its aggregation operation. Notably, {\em arch1} tends to duplicate the templates more often, which, thanks to the light-weight layers in each template, does not lead to a significant growth in the number of parameters.
 
\end{document}